\documentclass[10pt,twocolumn,letterpaper]{article}

\usepackage[pagenumbers]{cvpr}

\usepackage{graphicx}
\usepackage{amsmath}
\usepackage{amssymb}
\usepackage{booktabs}
\usepackage{cite}
\usepackage{times}
\usepackage{epsfig}
\usepackage{comment}
\usepackage{siunitx}
\usepackage{color}
\usepackage{colortbl}
\usepackage{xcolor}
\usepackage{multirow}
\usepackage{pifont}
\usepackage{enumitem}
\newcommand{\cmark}{\ding{51}}

\usepackage{algorithm}
\usepackage{listings}
\usepackage{url}

\newcommand{\algoreference}[1]{\textcolor{red}{}}

\usepackage[pagebackref,breaklinks,colorlinks]{hyperref}

\usepackage[capitalize]{cleveref}
\crefname{section}{Sec.}{Secs.}
\Crefname{section}{Section}{Sections}
\Crefname{table}{Table}{Tables}
\crefname{table}{Tab.}{Tabs.}

\begin{document}

\title{Transfer of Representations to Video Label Propagation:\\ Implementation Factors Matter}

\author{Daniel McKee$^1$\hspace{3mm}Zitong Zhan$^1$\hspace{3mm}Bing Shuai$^2$\hspace{3mm}Davide Modolo$^2$\hspace{3mm}Joseph Tighe$^2$\hspace{3mm}Svetlana Lazebnik$^1$ \\
$^1$University of Illinois at Urbana-Champaign, $^2$Amazon Web Services \\ 
{\tt\small \{dbmckee2,zitongz3,slazebni\}@illinois.edu, \{bshuai,dmodolo,tighej\}@amazon.com}
}

\maketitle

\begin{abstract}
  This work studies feature representations for dense label propagation in video, with a focus on recently proposed methods that learn video correspondence using self-supervised signals such as colorization or temporal cycle consistency. In the literature, these methods have been evaluated with an array of inconsistent settings, making it difficult to discern trends or compare performance fairly. Starting with a unified formulation of the label propagation algorithm that encompasses most existing variations, we systematically study the impact of important implementation factors in feature extraction and label propagation. Along the way, we report the accuracies of properly tuned supervised and unsupervised still image baselines, which are higher than those found in previous works. We also demonstrate that augmenting video-based correspondence cues with still-image-based ones can further improve performance. We then attempt a fair comparison of recent video-based methods on the DAVIS benchmark, showing convergence of best methods to performance levels near our strong ImageNet baseline, despite the usage of a variety of specialized video-based losses and training particulars. Additional comparisons on JHMDB and VIP datasets confirm the similar performance of current methods. We hope that this study will help to improve evaluation practices and better inform future research directions in temporal correspondence.
\end{abstract}

\section{Introduction}\label{sec:introduction}

\begin{figure}[!t]
\centering
\begin{tabular}{c}
\includegraphics[width=0.45\textwidth]{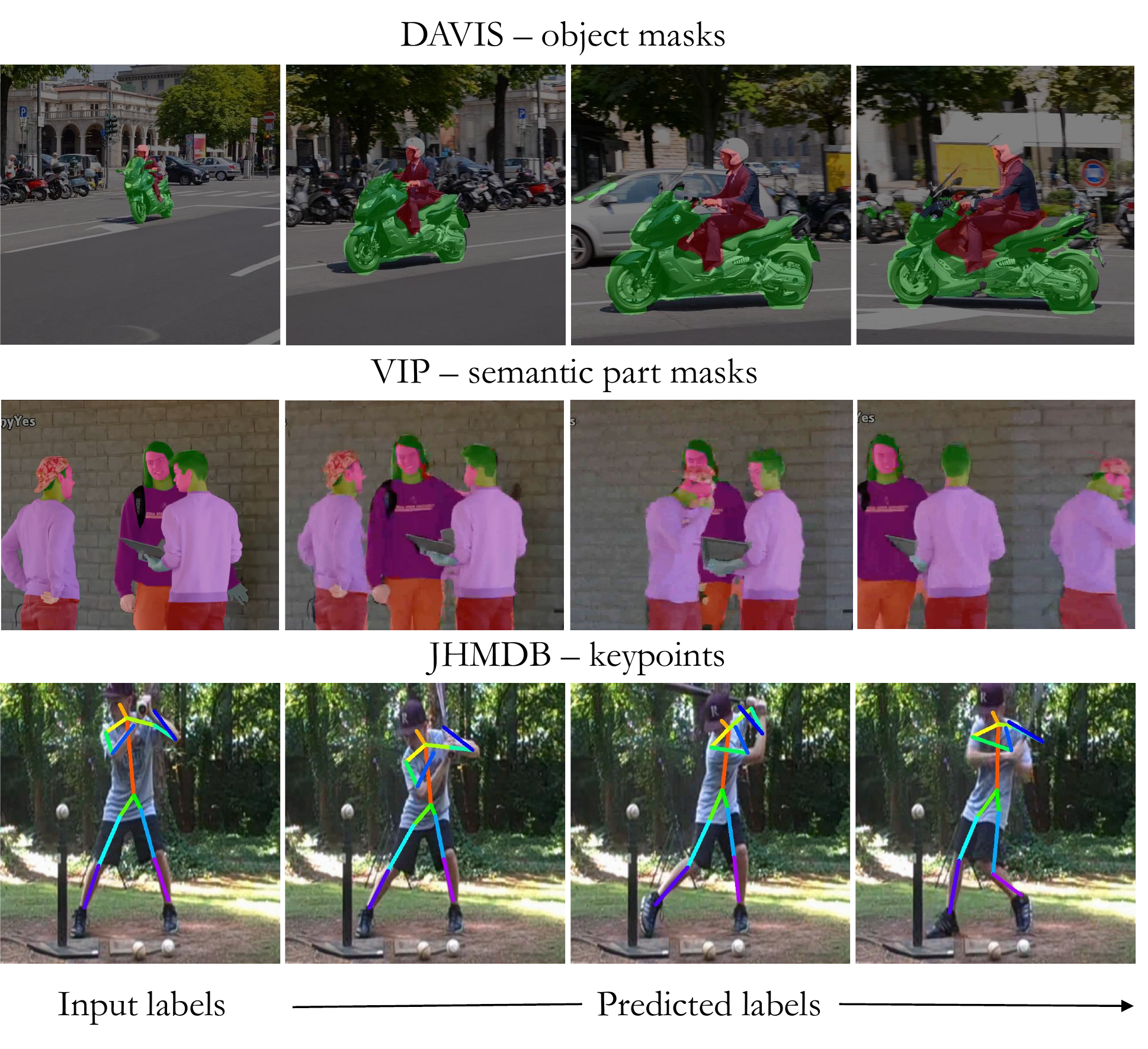} \\
\end{tabular}
\caption{Examples of video label propagation tasks and benchmarks: object masks on DAVIS~\cite{pont20172017} (top), semantic part masks on VIP~\cite{zhou2018adaptive} (middle), and human keypoints on JHMDB~\cite{jhuang2013towards} (bottom).}
\label{fig:teaser}
\end{figure}

Dense label propagation in video is a challenging task related to tracking and correspondence. Given a labeling in an initial frame, e.g., a segmentation mask or a set of keypoints, the goal is to propagate these labels forward through subsequent video frames (Figure \ref{fig:teaser}). Popular benchmarks for this task include DAVIS~\cite{pont20172017}, JHMDB~\cite{jhuang2013towards}, and VIP~\cite{zhou2018adaptive}. Obtaining the best accuracy on such benchmarks requires fully supervised training from large amounts of dense pixel annotations, often with an additional step of online adaptation to each test sequence~\cite{bao2018cnn,caelles2017one,luiten2018premvos,DMMNet19}. However, since large-scale pixel-level annotations are difficult to obtain, a number of recent works have focused instead on {\em self-supervised} approaches for learning temporal correspondence, where a feature representation pre-trained on a different dataset and pretext task is transferred to the desired label propagation benchmark with the help of a custom-designed propagation algorithm \cite{vondrick2018tracking,wang2019learning,lai2019self,li2019joint,lai2020mast,jabri2020walk,xu2021rethinking,araslanov2021dense}. 
Popular self-supervised pretext tasks include frame-to-frame colorization \cite{vondrick2018tracking, lai2020mast,lai2019self,li2019joint} and patch tracking with cycle consistency~\cite{wang2019learning,jabri2020walk}, while the propagation step relies on some variant of $k$-nearest-neighbors to label a location based on the most similar locations in previous frames, with the similarity computed from the pre-trained feature representation.

\begin{figure}
  \begin{center}
  \includegraphics[width=0.475\textwidth]{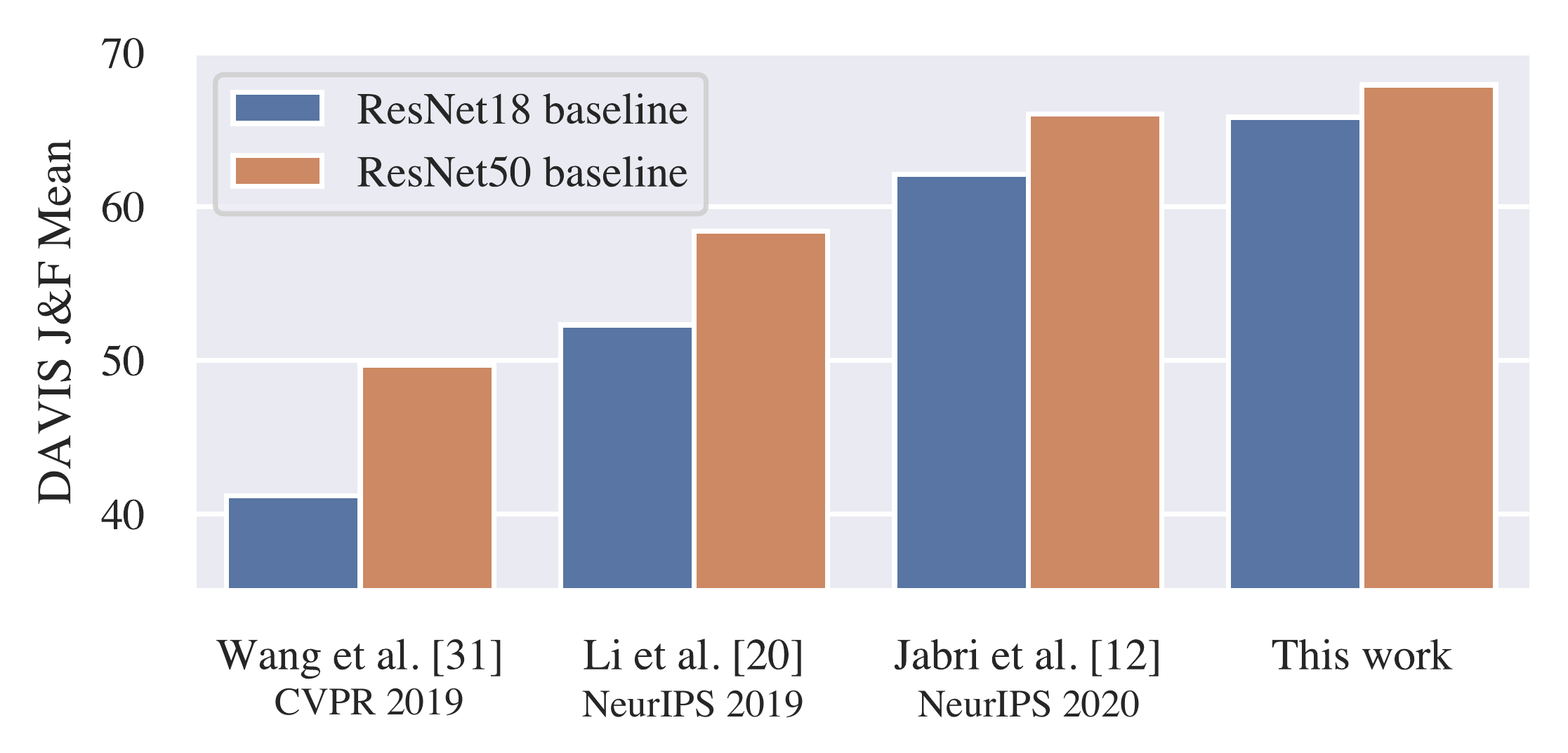}
  \caption{Comparison of reported performance for supervised ImageNet encoder baselines for label propagation on the DAVIS 2017 validation dataset from recent works.
  The three previous works rely on the same PyTorch pretrained ImageNet classification model, and the observed improvements are {\em solely related} to changes to evaluation settings and implementation for those works.\protect\footnotemark{} In this work, we get a further improvement by fine-tuning of late layers following stride modification, still solely on the ImageNet classification task (see Sec. \ref{sec:implementation-facts:feat-extract} for details).
  }
  \label{fig:imagenet-history}
  \end{center}
  \end{figure}

For understandable reasons, published papers on self-supervised label propagation tend to emphasize their pretext tasks and losses, and to downplay such mundane implementation factors as resolution of input video frames and output feature maps, the layer of the network from which the feature map is taken, the number of context frames in which nearest neighbors are found, and other hand-crafted details of the propagation algorithm. 
However, changing these factors can have at least as much influence on the performance of a method as the underlying feature representation.
To underscore this point, Figure \ref{fig:imagenet-history} shows the performance of a supervised ImageNet-pretrained baseline on DAVIS as reported by three recent papers~\cite{wang2019learning,li2019joint,jabri2020walk}. While each of these papers uses the exact same ImageNet-pretrained network, the reported performance varies dramatically due to differences in feature extraction and label propagation procedures (see caption and footnote for details). In our work, we show that the baseline accuracy can be improved even further (Figure \ref{fig:imagenet-history}, right) by fine-tuning the late layers following stride changes, but still {\em on the ImageNet classification task alone} (see Section \ref{sec:implementation-facts:feat-extract} for details). We argue that such attention to optimizing the baseline is justified because researchers tend to spend a lot of effort on getting every last bit of performance out of their own method -- often by tuning factors unrelated to their core technical contributions -- while neglecting to expend a similar amount of effort on the baseline. Establishing the strongest possible baseline is essential for conducting a clear-eyed evaluation of existing work. Beyond sometimes relying on suboptimal baseline numbers, papers in this area have compared their numbers to those of previously published methods employing a range of inconsistent settings, which makes it difficult to understand the causes of performance differences or to assess how much progress has really been made.
In the spirit of comparative studies that have taken a critical look at areas such as metric learning~\cite{musgrave2020metric}, self-supervised representation learning~\cite{kolesnikov2019revisiting}, and tracking~\cite{wang2021different}, we conduct a study to clarify performance differences between recent self-supervised video correspondence works by comparing several feature representations trained on still images and video. 

\footnotetext{Performance improvement in Li et al.\cite{li2019joint} arises from evaluating at full 480 pixel input resolution with preserved aspect ratio. Further improvements in Jabri et al. \cite{jabri2020walk} come from modifications to the affinity matrix computation (switching to the ``overall" approach) and addition of spatial context localization (see Sec. \ref{sec:propagation} for description of these factors). 
The ResNet18 result from TimeCycle and ResNet50 result from UVC are generated using released code.
}

We begin in Sec. \ref{sec:propagation} by presenting our unified label propagation algorithm to systematize the variations used in most existing works. Sec. \ref{section:approaches} summarizes the feature representations and video correspondence methods included in our study. Sec. \ref{sec:implementation-facts} evaluates the effect of several key implementation factors on our main benchmark, the DAVIS dataset. Here, we report the accuracies of properly tuned supervised and unsupervised still image baselines, which are higher than those found in previous works. Next, in Sec. \ref{sec:davis-comp:uvc-studies}, we investigate whether it is possible to further improve the performance of video-based correspondence models by augmenting their training with still image cues. We introduce a hybrid method relying on signals from both UVC~\cite{li2019joint} video-based learning and a self-supervised image objective. This method, which we call UVC+, outperforms both UVC and still-image-based methods, pointing in a promising future research direction. Next, Sec. \ref{sec:davis-comp} attempts a fair comparison of recent top-performing methods on the DAVIS dataset in light of our findings. Here, we find that several of the most recent methods, including our UVC+, CRW~\cite{jabri2020walk}, and VFS~\cite{xu2021rethinking}, achieve very similar levels of performance, close to our strong ImageNet baseline, despite relying on different combinations of cues and training particulars. Lastly, Sec. \ref{sec:vip-jhmdb} presents similar evaluations of existing methods on the JHMDB and VIP datasets. Finally, Sec. \ref{sec:conclusion} closes the paper with recommendations for future best practices.

\section{Generic Propagation Algorithm}\label{sec:propagation}

\begin{figure}[b]
  \centering
  \includegraphics[width=0.475\textwidth]{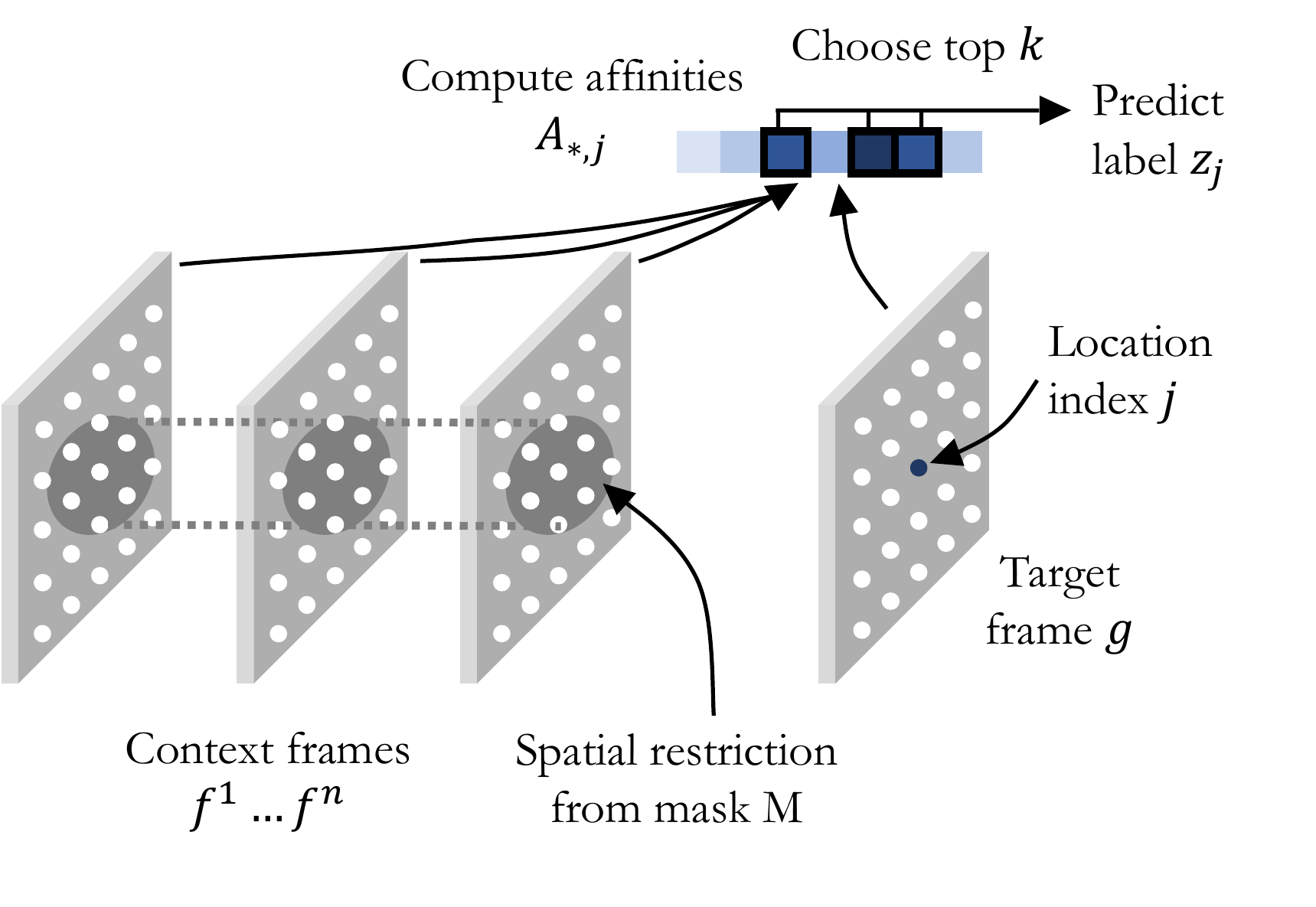}
  \caption{\textbf{A generic framework for label propagation.} 
  Given a  location in the target frame $g$, similarity is computed with each location in previous context frames $f^1,\ldots,f^n$. The context locations with highest similarity contribute to the label prediction for the target location. Spatial masking may be employed to restrict influence to the most relevant locations in context frames. Illustrated here is the ``overall" approach which aggregates label predictions across all context frames together.
  }
  \label{fig:prop}
  \end{figure}

As stated above, published works on self-supervised video correspondence learning adopt an inconsistent array of hand-crafted label propagation procedures without clearly describing all the relevant details, so recognizing important discrepancies can be difficult. We therefore start by formulating a reasonably unified ``umbrella'' propagation algorithm to systematize differences between works. Figure \ref{fig:prop} illustrates our setup and Algorithm 1 gives a PyTorch-like pseudocode summary.

\begin{algorithm}[t]
\caption{Propagation algorithm pseudocode (see text).}
\label{alg:prop}
\definecolor{codeblue}{rgb}{0.25,0.5,0.5}
\definecolor{codekw}{rgb}{0.85, 0.18, 0.50}
\lstset{
  backgroundcolor=\color{white},
  basicstyle=\fontsize{7.5pt}{7.5pt}\ttfamily\selectfont,
  columns=fullflexible,
  numbers=left,
  breaklines=true,
  xleftmargin=.7cm,
  captionpos=b,
  commentstyle=\fontsize{7.5pt}{7.5pt}\color{codeblue},
  keywordstyle=\fontsize{7.5pt}{7.5pt}\color{codekw},
}
\begin{lstlisting}[language=python]
# g: Target frame features 
# F: Context frame features F={f1,...,fn}
# Y: Labels Y={y1,...,yn}
# overall: Boolean for aggregation method
# k: Number of neighbors
# T: Softmax temperature
# z: Target frame label prediction

z = zeros_like(Y[0]) # L x N
if overall:
    # output dimensions C x (N x n),  L x (N x n)
    F, Y = [concat(F)], [concat(Y)]

# for each context feature matrix and label matrix
for f, y in zip(F,Y):
    # dimension P = N x n if overall, N otherwise
    A = compute_affinity(f,g) # P x N
    # mask generated for each class label restricts context and/or target regions
    M = compute_masks(F,Y,g)  # L x P x N
    
    # for each label l, apply mask and compute topk
    for l in range(len(M)):
        A_l = M[l, :,:] * A
        # for each target location
        for j in range(N):
            # top k affinities and indices
            topk_vals, topk_inds = topk(A_l[:,j])
            # gather label vectors for top k indices
            topk_labels = y[l, topk_inds]
            if overall:
                topk_vals = softmax(topk_vals/T)
            # predict target label from weighted top k context labels
            z[l,j] += topk_vals * topk_labels

# average predictions over context frames
z=1/len(F)*z

def compute_affinity(f1, f2):
    aff = mm(f1.T, f2)
    if not overall:
        aff = softmax(aff/T, dim=1)
    return aff


\end{lstlisting}
\end{algorithm}

\textbf{Setup and feature extraction.} At a given time step, let
$f^1, ... , f^n$ denote {\em context frames} with corresponding label vectors $y^1, ..., y^n$ and $g$ denote the {\em target frame} to which the context labels must be propagated to predict the target label vector $z$. Concretely, all the frames are represented by flattened feature vectors produced by a feature extraction network, so their dimensions are $C \times N$, where $C$ is the number of feature channels output by the network and $N=W \cdot H$ is the product of the spatial dimensions of the feature map. The dimensions of the label vectors are $L \times N$, where $L$ is the number of semantic classes. At each location, the label vector is one-hot in reference annotations, but can take on values between 0 and 1 at inference time.

\textbf{Affinity computation.} The first step is to compute affinities between locations in the target and context frames. Letting $f_i$ and $g_j$ denote the $i$th and $j$th columns in $f$ and $g$, representing the feature vectors at the respective locations, we define

\begin{equation}
   A_{i j}= {f_{i}}^{T} g_{j} \,.
   \label{eq:aff}
\end{equation}

Among existing approaches, there are two variants for computing the affinities, which we refer to as ``overall'' and ``per-frame''. In the ``overall'' variant, adopted by CRW~\cite{jabri2020walk}, a single affinity matrix is computed between each target location and all context frame locations simultaneously. In Algorithm 1, this is done by a $\verb|concat|$ pre-processing operation (lines 10-11) to transform the $n$ context features $f^{1}, ..., f^{n}$ each of dimension ${C \times N}$ into a single feature matrix of dimension ${C \times (N \cdot n)}$, and the labels $y^1, ..., y^n$ of dimension ${L \times N}$ to a matrix of dimension $L \times (N \cdot n)$. In the ``per-frame'' variant, adopted by most of the other approaches, $n$ separate affinity matrices are computed between the target and each context frame. Softmax normalization is applied differently in these variants, as explained below.

\textbf{Spatial context masking.} Many methods heuristically restrict the sets of context locations that are propagated {\em from} and target locations that are propagated {\em to}. All such restrictions can be accomplished by element-wise multiplication of the affinity matrix with a binary mask $M_{\ell}$. In particular, the CRW method~\cite{jabri2020walk} only propagates labels from context frame regions within some radius of the target location. To do this, for a target location (column index) $j$ in the affinity matrix, we mask out all context locations (row indices) $i$ that lie outside the specified radius. UVC~\cite{li2019joint}, the other major method we evaluate, uses a more complex tracking scheme that restricts the propagation target region for a particular class based on the regions occupied by that class in prior frames. This requires class-specific masks, as reflected in line 23\algoreference{} of Algorithm 1.

\textbf{Top $k$ aggregation.} To predict the value $z_{\ell j}$ for label $\ell$ at target frame location $j$, we find the set of $k$ context indices with the highest (masked) affinities to that location, and then perform a ``soft copy'' operation (line 33) \algoreference{} in which the target label value is set to the sum of context label values at the top $k$ locations weighted by their respective normalized affinities (see below). For the ``per-frame'' approach, this operation is performed separately w.r.t. each context frame, and the resulting label values are finally averaged (line 36). \algoreference{}

\textbf{Softmax normalization.} For the ``overall'' approach, softmax normalization happens after the top $k$ affinity values are selected as described above (line 31). \algoreference{} For the ``per-frame'' approach, the affinity matrix for each frame is normalized column-wise, i.e., the affinity values of that frame w.r.t. each target location $j$ sum to one. In Algorithm 1, this is done in line 41,\algoreference{} {\em before} the masking step, which is consistent with the UVC approach; alternatively, this can be done {\em after} the masking, as in CorrFlow and MAST.\footnote{With softmax normalization done before the masking and top-$k$ selection, one might worry about the class label values decaying over time, since the top-$k$ weights will not add to one. In practice, however, this does not seem to pose a problem. With the UVC method on the DAVIS dataset, we have observed the affinity matrix values to be quite peaky initially, so that the average predicted label vector sums stay in the range $0.25-0.75$ through the end of the DAVIS videos, which have around 50-80 frames to propagate.} In all cases, a temperature parameter $T$ is used to determine how ``sharp'' the softmax values become.
\smallskip

To summarize, the implementation choices in the above-described propagation procedure that can affect final performance include the following:
\begin{itemize}[noitemsep]
\item Affinity computation and label aggregation approach (overall vs. per-frame).
\item Softmax temperature $T$.
\item Number of top neighbor locations $k$.
\item Number of context frames $n$.
\item Spatial restriction and masking.
\end{itemize}
In Section \ref{sec:implementation-facts}, we will examine the impact of these choices.

\section{Feature Representations}\label{section:approaches}

In this section, we introduce the still image feature representation baselines and self-supervised temporal correspondence approaches included in our study. Sec. \ref{section:approaches:image} describes our baselines, which include a supervised network trained on ImageNet classification, and self-supervised networks trained using rotation prediction~\cite{gidaris2018unsupervised} and the more recent Momentum Contrast (MoCo) method~\cite{he2020momentum}. Sec. \ref{sec:correspondence_methods} describes the video correspondence methods we study, which include TimeCycle~\cite{wang2019learning}, CorrFlow~\cite{lai2019self}, MAST~\cite{lai2020mast}, UVC~\cite{li2019joint}, and CRW~\cite{jabri2020walk}. These are among the most prominent and representative methods from the last few years. 

\subsection{Still Image Baselines} \label{section:approaches:image}

The generic label propagation algorithm presented in the previous section can be applied on top of features produced by any network -- supervised or unsupervised, trained on still images or video. Therefore, it makes sense to start with a few well-known still image networks as baselines. Unless otherwise specified, all our networks use the ResNet18 architecture.

\textbf{Supervised ImageNet.} 
Despite being trained with only image-level labels and no video data, a ResNet trained on ImageNet classification sets a strong baseline for the temporal correspondence tasks~\cite{wang2019learning}. The fact that ImageNet-trained features are useful for correspondence is well documented. As early as 2014, Long et al.~\cite{long2014convnets} previously showed that features localize at a much finer scale than their receptive field size and perform well on semantic image alignment tasks. 

\textbf{Self-supervised Rotation and MoCo.} The quality of features learned by recent self-supervised methods has approached or even surpassed that of supervised ones for segmentation and detection tasks \cite{he2020momentum,misra2020self,chen2020simple,chen2021exploring}. We chose two representative self-supervised methods for our experiments. The first is the rotation prediction method of Gidaris et al.~\cite{gidaris2018unsupervised}, which is exceedingly straightforward and effective in many settings as an auxiliary loss \cite{zhai2019s4l,chen2019self,gidaris2019boosting,misra2020self}. The second is Momentum Contrast (MoCo)~\cite{he2020momentum}, which comes from the more recent family of contrastive methods \cite{he2020momentum,chen2020simple,misra2020self} that have achieved top performance when transferred to downstream image understanding tasks.

\subsection{Video-Based Correspondence Methods} \label{sec:correspondence_methods}

Self-supervised video correspondence methods define training tasks utilizing an affinity matrix $A^{a\rightarrow b}$ between a target frame $b$ and context frames $a$, similar to the propagation process described in Sec. \ref{sec:propagation}. One of the first such methods was introduced by Vondrick et al.~\cite{vondrick2018tracking}, which used the affinity matrix to copy quantized color labels from reference frames to a grayscale target frame. Numerous follow-ups have since been released building on and refining such mechanisms. Below, we summarize the video correspondence methods we evaluate, including TimeCycle~\cite{wang2019learning}, CorrFlow~\cite{lai2019self}, MAST~\cite{lai2020mast}, UVC~\cite{li2019joint}, and CRW~\cite{jabri2020walk}.

\vspace{.5em}
\textbf{TimeCycle.} In place of colorization, the TimeCycle method of Wang et al.~\cite{wang2019learning} proposes temporal cycle consistency between frames as a supervision method. Specifically, TimeCycle computes an affinity matrix $A^{p_1\rightarrow I_2}$ between features from a random patch $p_1$ cropped from frame $I_1$ and a second frame $I_2$. Based on the affinity matrix, it employs a spatial transformer module~\cite{jaderberg2015spatial} to predict the location of the $p_1$ patch in $I_2$ and applies differentiable cropping to obtain $\hat{p_2}$.
After similarly computing an affinity matrix $A^{\hat{p}_2  \rightarrow I_1}$ and predicting a patch $\hat{p}_1$, it uses the differences in both position and feature representation of $p_1$ and $\hat{p}_1$ as supervisory signals. TimeCycle applies this mechanism recurrently to track forward and backward through longer sequences. During evaluation, this method uses a per-frame propagation approach with no spatial context masking.

\vspace{.5em}
\textbf{UVC.} The UVC approach of Li et al.~\cite{li2019joint} combines colorization with a higher-level patch tracking task and temporal cycle consistency loss similar to TimeCycle.
Unlike TimeCycle, UVC does not use an auxiliary spatial transformer network to predict patch location parameters. Instead, an $(x,y)$ coordinate location for patch $p_2$ can be estimated directly by using the affinity matrix $A^{I_2\rightarrow p_1}$. A grid coordinate prediction for location $j$ from $p_1$ in $I_2$ is computed as $(\sum_{i} A^{I_2\rightarrow p_1}_{ij} u_i^{I_2},    \sum_{i} A^{I_2\rightarrow p_1}_{ij} v_i^{I_2})$ where $(u_i^{I_2}, v_i^{I_2})$ is the grid coordinate of location $i$ in $I_2$. By aggregating estimated coordinates for all locations in $p_1$, this method infers a coordinate center for $p_2$. Conveniently, it may also be used to predict an ROI box during evaluation for object mask propagation (referred to as ``track'' inference).
UVC is trained with a combination of three losses: a colorization loss, a cycle consistency loss, and a concentration loss. Colorization loss is an L1 loss over predicted colors in Lab space, which shows substantial improvement over the quantized color cross-entropy loss of Vondrick et al.~\cite{vondrick2018tracking}. Specifically, the affinity matrix $A^{p_1 \rightarrow \hat{p}_2}$ between patch $p_1$ and patch $\hat{p}_2$ is used to copy features which are encoded and decoded into Lab color space using a pretrained autoencoder. The cycle consistency loss is posed as an L1 penalty on reconstructed color for $p_1$, which is shown to be equivalent to orthogonal regularization of the $A^{p_1 \rightarrow \hat{p}_2}$ matrix. 
Lastly, the concentration term penalizes patches of locations that do not move together by applying an L2 loss for each grid coordinate within a patch from its center coordinate. For evaluation, UVC uses the per-frame propagation approach together with the ``track'' technique mentioned above. Since the affinity matrix computation is included in the training pipeline, the test-time behavior of UVC can be very sensitive to propagation algorithm implementation changes as discussed in Sec. \ref{sec:implementation-facts}.

\vspace{.5em}
\textbf{CorrFlow and MAST.} Correspondence Flow (CorrFlow) by Lai and Xie~\cite{lai2019self} and follow-up method Memory-Augmented Self-supervised Tracker (MAST) by Lai et al.~\cite{lai2020mast} are both colorization-based approaches. CorrFlow incorporates a temporal cycle consistency constraint similar to UVC, while MAST relies purely on improved colorization training. These methods both use color channel dropout rather than grayscale input frames to improve performance, and MAST shows that a dropout in Lab rather than RGB space is substantially more effective. CorrFlow uses a cross-entropy loss over quantized color classes, but MAST shows that a regression loss as used by UVC is more effective. MAST also incorporates a more sophisticated memory-based attention mechanism during training and inference to predict spatial context. 
Both CorrFlow and MAST propagate labels using the per-frame approach with one difference: softmax normalization is applied only to the square region of interest (ROI) selected by spatial context masking in each context frame. 
The center of the ROI is fixed at the target location coordinate for CorrFlow, but MAST selects the ROI center for a context frame through a separate sparse affinity matrix computation. 

\vspace{.5em}
\textbf{CRW.} The Contrastive Random Walk (CRW) method by Jabri et al.~\cite{jabri2020walk} follows TimeCycle in using temporal cycle consistency as a primary supervision signal. 
In this case, patches are cropped from a grid of locations in a frame and separately encoded into feature representations to prevent learning shortcut solutions. 
A single feature vector representation is pooled from each patch, and affinities are computed between all patch representations in a frame $a$ and all patch representations for a subsequent frame $b$. 
The training task is posed as a probabilistic random walk following the affinities between patches forward and backward through time. This work makes several improvements over TimeCycle including probabilistically rather than deterministically tracking the paths of patches and simultaneously tracking many patches through time. 
As a result, unlike TimeCycle, CRW benefits from tracking longer range sequences and attains stronger object-level features using a technique which drops out patch correspondence paths during training.
During CRW's label propagation, the affinity matrix computation is different from training as features are computed for the full frame rather than individual patches and there is no pooling applied over these representations. Further, since CRW relies on the ``overall" approach, a single affinity matrix is computed between the target frame and all context frames at test time (rather than between patches in only consecutive frames like training), and spatial context masking is applied only at inference to restrict context regions. 

\vspace{.5em}
\textbf{VFS.} The Video Frame-level Similarity (VFS) approach \cite{xu2021rethinking} applies image-based similarity learning based on SimSiam~\cite{chen2021exploring} to a video setting. 
This method does not rely on any specialized tracking or correspondence mechanisms. Instead, VFS trains with only an image-level similarity loss over sampled video frame pairs: any two pairs of frames sampled from the same video should have a high similarity according to the learned feature representation. The video frame pair sampling can be considered as a form of data augmentation which is added on top of typical image-based augmentations like random crops, flips, and color jitter from SimSiam.
The learned features that emerge from this learning process reach performance comparable to top self-supervised correspondence methods. Similarly to CRW, this work relies on the overall propagation approach and applies spatial context masking of a fixed radius.

\subsection{Training Details} \label{sec:training_details}

In the following experiments, we rely both on previously released models and ones we train. Our self-supervised still image baselines are trained from scratch using the settings below. For TimeCycle and CRW, we use provided evaluation code and report the results. For UVC, we train our own models based on released code. For CorrFlow, MAST, and VFS, we cite reported results directly.

\textbf{Image-pretrained baselines.} To be consistent with recent works~\cite{li2019joint,lai2020mast,jabri2020walk}, we select ResNet18
as the architecture for our baselines. For the ImageNet classification baseline, we use the pre-trained PyTorch ResNet18 model~\cite{paszke2019pytorch}. For rotation prediction, we train a ResNet18 from scratch on ImageNet~\cite{deng2009imagenet} for 180 epochs using SGD with weight decay of 0.0001 and momentum of 0.9. An initial learning rate of 0.1 is decayed by a factor of 10 every 30 epochs. We set batch size to 256. For MoCo training, we use default settings from the paper and released code.

For the image-based models fine-tuned with stride change, as described in Sec. \ref{sec:implementation-facts:feat-extract}, we modify the stride in \verb|res3| and fine-tune only \verb|res3| and later layers, freezing all earlier layers. For these models, we fine-tune for 40 epochs on the original training task. We set initial learning rates of 0.01 for ImageNet classification and Rotation and 0.003 for MoCo. We decay the learning rate by a factor of 10 after 20 epochs.

\textbf{UVC.} We use training settings consistent with the UVC released model. However, we find that the UVC model benefits from additional training, so we train for a total of 50 epochs on the Kinetics dataset~\cite{kay2017kinetics}.

\section{Label Propagation Studies on DAVIS}
\label{sec:implementation-facts}

In this section, we evaluate key implementation factors on our
primary benchmark, the widely used DAVIS video object segmentation dataset~\cite{pont20172017}. Sec. \ref{sec:davis} will discuss this benchmark and its evaluation protocol, Sec. \ref{sec:implementation-facts:feat-extract} will consider implementation factors related to feature extraction, and Sec. \ref{sec:implementation-facts:prop} will focus on those related to label propagation.

\subsection{DAVIS Dataset} \label{sec:davis}

The DAVIS 2017 dataset~\cite{pont20172017} consists of short video sequences of around 70 frames annotated with pixel-level object masks. 
It is split into training, validation, and two test sets. The training set consists of 60 sequences, while each of the validation and test sets consists of 30 sequences. Evaluation for the two test sets, test-dev and test-challenge, is housed on a submission server. Access to the test-challenge dataset has been restricted to a limited number of submissions during active DAVIS Challenge periods, while the test-dev server continues to allow regular submissions. 
For ease of evaluation, all the previous self-supervised correspondence works we consider~\cite{wang2019learning,lai2019self,lai2020mast,li2019joint,jabri2020walk}
report results on the validation set, rather than evaluating on the test server. Notably, all these methods have also tuned their propagation implementation on the same validation set. In order to be able to trust highly-tuned performance numbers on the validation set, the tuning process needs to be made more transparent and reproducible by others. Accordingly, our work attempts to identify all the relevant hyperparameters and report results for different combinations of values. For future evaluations, it would be better for researchers to also report results on a separate test set.

 For evaluation of accuracy on DAVIS, we report the standard $\mathcal{J}$ and $\mathcal{F}$ metrics~\cite{perazzi2016benchmark}. The $\mathcal{J}$ (Jaccard index) metric measures the intersection over union of the predicted object mask with the ground truth mask. The $\mathcal{F}$ metric measures the boundary accuracy by finding the cost of a bipartite matching between boundary pixels of ground truth and predicted masks.
 For these metrics, we compute the mean ($\mathcal{M}$) of scores across all sequences and the recall ($\mathcal{O}$) which is the fraction of sequences scoring over a fixed threshold (set to 0.5), as is standard.
 In some cases, we also report $\mathcal{J}\&\mathcal{F}_{\mathcal{M}}$, the mean of $\mathcal{J}_{\mathcal{M}}$ and $\mathcal{F}_{\mathcal{M}}$.

\subsection{Feature Extraction} \label{sec:implementation-facts:feat-extract}

As explained earlier, self-supervised label propagation methods consist of a pre-trained feature extraction stage followed by a separate, hand-crafted propagation algorithm. In this section, we identify and evaluate important implementation factors for the feature extraction step. For all experiments presented in this subsection, we use per-frame evaluation with nearest neighbors $k=5$ and number of context frames $n=7$.

\setlength{\tabcolsep}{4pt}
\begin{table}
\begin{center}
\caption{Performance on DAVIS at different input resolutions for ImageNet and UVC models.
In these experiments, the models are unchanged. Note how increasing the resolution causes dramatic increase in performance.}
\label{table:davis-res}
{\small
\begin{tabular}
{p{0.05\textwidth}>{\centering}p{0.13\textwidth}>{\centering}p{0.05\textwidth}>{\centering}p{0.05\textwidth}>{\centering}p{0.05\textwidth}>{\centering\arraybackslash}p{0.05\textwidth}}
\toprule 
Model & Resolution & $\mathcal{J}_{\mathcal{M}}$ & $\mathcal{J}_{\mathcal{O}}$ & $\mathcal{F}_{\mathcal{M}}$ & $\mathcal{F}_{\mathcal{O}}$\\
\midrule
ImageNet & 320$\times$320  & 46.6 & 48.9 & 47.0 & 48.3 \\
& 480$\times$480 & 48.6 & 50.9 & 54.2 & 56.1 \\
& 480$\times$full & \textbf{51.3} & \textbf{56.7}  & \textbf{56.6} & \textbf{59.6} \\
\midrule 
UVC & 320$\times$320        & 46.7 & 49.0  & 44.6 & 42.1  \\
& 480$\times$480        & 52.5 & 60.3  & 54.3 & 57.3  \\
& 480$\times$full  & \textbf{56.3} & \textbf{64.9}  & \textbf{59.2} & \textbf{64.0} \\
\bottomrule 
\end{tabular}
}
\end{center}
\end{table}
\setlength{\tabcolsep}{1.4pt}

\textbf{Input frame resolution.} Some of the earlier label propagation works, notably Wang et al.~\cite{wang2019learning}, downsampled video frames to 320$\times$320 resolution, while more recent works (UVC~\cite{li2019joint}, CorrFlow~\cite{lai2019self}, MAST~\cite{lai2020mast}, and CRW~\cite{jabri2020walk}) use full resolution with preserved aspect ratio (most DAVIS videos are $480 \times 854$, but widths can vary). Table \ref{table:davis-res} shows performance of our supervised ImageNet baseline and the UVC released model evaluated at multiple resolution settings. The varying input resolutions are downsampled by our fully convolutional networks by a factor of 1/8 (see below) to produce feature grids of varying sizes: in particular, $480 \times 480$ frames give $60 \times 60$ grids, and $320 \times 320$ frames give $40 \times 40$ grids. Class scores are then predicted at the resolution of the feature grid, upsampled to original video resolution, and finally converted to label maps to evaluate against the ground truth masks. Smaller input resolutions and feature grids require additional upsampling, which can harm the sharpness and accuracy of predicted masks. Table \ref{table:davis-res} confirms that the lowest setting, consistent with TimeCycle~\cite{wang2019learning}, significantly limits performance, as does resampling to a square aspect ratio. Overall, resolution changes alone can cause an over 10-point difference in performance, as we can see with the UVC model. This is important to keep in mind since some papers simply list earlier published results, like those of TimeCycle,  without explicitly noting the resolution discrepancy.

\setlength{\tabcolsep}{4pt}
\begin{table}
\begin{center}
\caption{
Performance of DeepCluster VGG16 and ImageNet ResNet18 networks using different feature extraction layers. For DeepCluster, we include the previously reported result (top) as well as performance of features from conv10 and conv13 using our evaluation code. 
The choice of feature extraction layer makes a substantial performance difference.
}
\label{table:feat-layer}
{\small
\begin{tabular}
{p{0.125\textwidth}>{\centering}p{0.05\textwidth}>{\centering}p{0.07\textwidth}>{\centering}p{0.03\textwidth}>{\centering}p{0.03\textwidth}>{\centering}p{0.03\textwidth}>{\centering\arraybackslash}p{0.03\textwidth}}
\toprule 
 Method & Layer & Resolution & $\mathcal{J}_{\mathcal{M}}$ & $\mathcal{J}_{\mathcal{O}}$ & $\mathcal{F}_{\mathcal{M}}$ & $\mathcal{F}_{\mathcal{O}}$\\
\midrule
DeepCluster \cite{wang2019learning} &  \footnotesize{conv13} & \footnotesize{320$\times$320} &  37.5 & - & 33.2 & - \\
DeepCluster  & \footnotesize{conv13} &  \footnotesize{320$\times$320} &  38.8&	34.2& 	34.6&	21.8 \\
                   & \footnotesize{conv10} &   \footnotesize{320$\times$320} &  46.6	& 49.0	&	45.7 &	46.3\\
                   & \footnotesize{conv10} &   \footnotesize{480$\times$full} & \textbf{48.2} & \textbf{49.2}  & \textbf{53.2} & \textbf{53.8}  \\
\midrule 
ImageNet & \footnotesize{res2} &   \footnotesize{480$\times$full} &  37.8 & 35.8 & 45.2 & 42.6  \\ 
&  \footnotesize{res3} & \footnotesize{480$\times$full} & \textbf{51.3} & \textbf{56.7}  & \textbf{56.6} & \textbf{59.6} \\
& \footnotesize{res4} &  \footnotesize{480$\times$full} & 42.2 & 41.2 & 44.6 & 43.6  \\ 
\bottomrule  
\end{tabular}
}
\end{center}
\end{table}
\setlength{\tabcolsep}{1.4pt}

\textbf{Network downsampling factor.} Most propagation methods, including UVC, TimeCycle, and CRW, use feature maps that are downsampled by a factor of 1/8 from the original frames, and we stick with the same setting in our experiments. However, some of the more recent methods, notably CorrFlow and MAST, obtain a smaller downsampling factor of 1/4 by removing the maxpooling layer in ResNet18. We do not go to the expense of training alternative models with this modification, but in Sec. \ref{sec:davis-comp}, Table \ref{table:davis-main}, we will evaluate our modified version of UVC, as well as CRW, with upsampled feature grids and show that this can lead to an improvement even without re-training.

\textbf{Feature extraction layer.} Features in different layers differ in their spatial selectivity and expressiveness, so choosing the best layer is important. Note, to maintain the same downsampling factor when switching feature extraction layers, it may also be necessary to modify stride or remove maxpool layers. As an example of where feature extraction can make a drastic difference in previously reported baselines, we revisit the performance of a DeepCluster~\cite{caron2018deep} VGG16 network used by Wang et al.~\cite{wang2019learning} as a self-supervised image-trained baseline. The top line of Table \ref{table:feat-layer} lists the published result from Table 1 of~\cite{wang2019learning}, which was obtained by extracting features from \verb|conv13| after removing the \verb|maxpool4| layer after \verb|conv10|, giving a 1/8 downsampling factor. The second line of Table \ref{table:feat-layer} reproduces the evaluation using our code (primarily inherited from UVC). As an alternative to \verb|conv13| features, we can maintain the same downsampling factor by extracting features from \verb|conv10| without removing any maxpool layers. As a result of this change, as shown in the third line of Table \ref{table:feat-layer}, $\mathcal{J_M}$ dramatically improves from 38.8 to 46.6, and $\mathcal{F_M}$ improves from 34.6 to 45.7. This would have been enough to change the conclusions of the TimeCycle work, since the improved DeepCluster baseline substantially outperforms the reported TimeCycle numbers ($\mathcal{J_M}$ of 40.1 and $\mathcal{F_M}$ of 38.3 for ResNet18, from Table 1 of~\cite{wang2019learning}). Processing frames at the full resolution, as in the fourth line of our Table \ref{table:feat-layer}, yields further increases, particularly in the $\mathcal{F}$ metrics.

In the rest of our experiments, we use ResNet18 instead of VGG16, but similar considerations apply.
In the lower half of Table \ref{table:feat-layer}, we report performance using features from \verb|res2|, \verb|res3| and \verb|res4| layers for our supervised ImageNet model. To maintain a 1/8 downsampling factor for \verb|res3| and \verb|res4|, their strides must be changed from 2 to 1. We can see that $\verb|res3|$ features perform much better than the early $\verb|res2|$ features, or the later $\verb|res4|$ ones. For the subsequent experiments, we stick with the \verb|res3| features (with stride change), which is consistent with previous temporal correspondence methods such as UVC.

\textbf{Fine-tuning after stride modification.} 
Changing the stride in a layer of a pre-trained network changes the receptive field for subsequent layers and could degrade feature quality. To overcome any adverse effects for our still image networks, we  fine-tune \verb|res3| and later layers with the original objective (ImageNet classification, rotation prediction~\cite{gidaris2018unsupervised}, or MoCo~\cite{he2020momentum}) after changing the \verb|res3| stride to 1 (see Section \ref{sec:training_details} above for training details).
As shown in Table \ref{table:arch}, this fine-tuning gives increases of around 2-3 points in $\mathcal{J}$ and $\mathcal{F}$ metrics. Thus, in subsequent experiments, we use the fine-tuned image-based models.

\setlength{\tabcolsep}{4pt}
\begin{table}
\begin{center}
\caption{Comparison of performance of image-based networks on DAVIS before and after fine-tuning layers with modified stride. 
Fine-tuning consistently improves performance across methods.}
\label{table:arch}
{\small
\begin{tabular}
{p{0.08\textwidth}>{\centering}p{0.1\textwidth}>{\centering}p{0.05\textwidth}>{\centering}p{0.05\textwidth}>{\centering}p{0.05\textwidth}>{\centering\arraybackslash}p{0.05\textwidth}}
\toprule
 Method & Fine-tuned & $\mathcal{J}_{\mathcal{M}}$ & $\mathcal{J}_{\mathcal{O}}$ & $\mathcal{F}_{\mathcal{M}}$ & $\mathcal{F}_{\mathcal{O}}$\\
\midrule
ImageNet  & \textbf{-} & 51.3 & 56.7  & 56.6 & 59.6     \\
 &\cmark & \textbf{53.4} & \textbf{58.8} &	\textbf{59.0} & \textbf{63.0} \\
\midrule
Rotation & \textbf{-}  & 44.8	& 45.8 & 	50.7 & 	50.1 \\
 &\cmark &   \textbf{48.1}	& \textbf{50.1} &	\textbf{54.3}	& \textbf{54.8} \\
\midrule
MoCo & \textbf{-} &  51.8	  & 57.3	& 56.4 &	59.3 \\
 & \cmark & \textbf{53.2} & 	\textbf{59.1}  & 	\textbf{57.5}	 & \textbf{60.2} \\
\bottomrule 
\end{tabular}
}
\end{center}
\end{table}
\setlength{\tabcolsep}{1.4pt}

\textbf{Feature vector normalization.} Also worth mentioning, though not explicitly evaluated here, is feature vector normalization. For the still image baselines and CRW, we found L2 normalization of the feature vectors to be very important. On the other hand, UVC, MAST, and CorrFlow do not normalize feature vectors, and adding normalization at test time would likely hurt performance since these models are trained without it.

\begin{figure*}
\begin{center}
\caption{Plots of DAVIS evaluation under varied hyperparameter settings for selected models. Performance is plotted as the averaged $\mathcal{J}\& \mathcal{F}_{\mathcal{M}}$ metric.
We provide a study of the softmax normalization temperature parameter as well as context and nearest neighbors for per-frame and overall propagation approaches. For experiments in plots (b) and (c), we use the best temperatures for each model from plot (a). For overall: 1/Temp is set to 20 for CRW and 1 for all others. For per-frame: 1/Temp is set to 40 for ImageNet and Rotation, 50 for MoCo and CRW, and 1 for UVC. We fix number of nearest neighbors $k=5$ in plots (a) and (b), and we fix number of context frames $n=7$ in plots (a) and (c). 
}
\label{table:hyperparameter-studies}
\includegraphics[width = 7.5in]{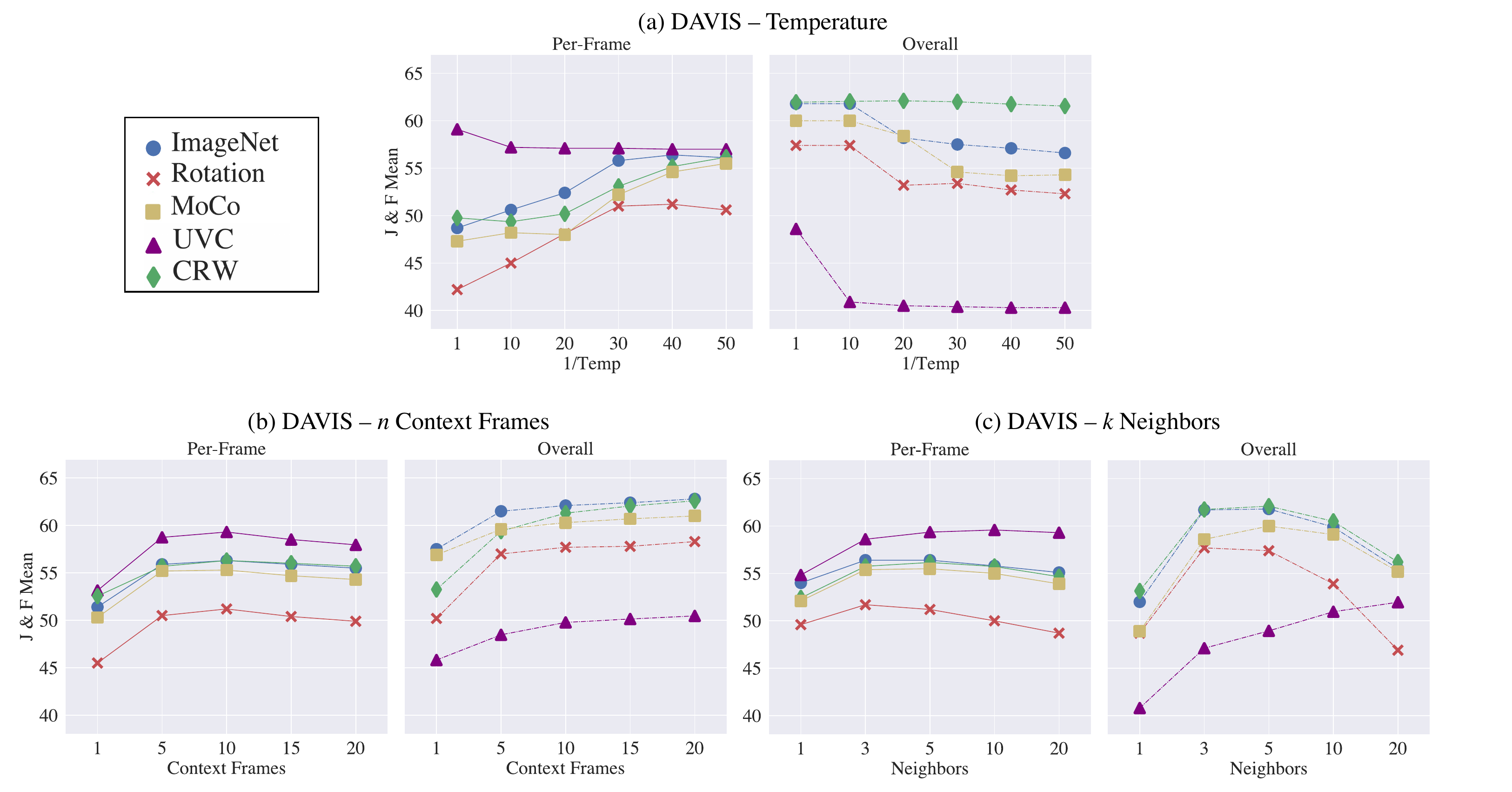}
\end{center}
\end{figure*}

\subsection{Propagation Algorithm}\label{sec:implementation-facts:prop}

Next, we examine the effect of implementation choices in the propagation algorithm that were identified in Sec. \ref{sec:propagation}.

\textbf{Overall vs. per-frame aggregation.}  A key choice is the means of aggregating labels over context frames. As stated in Section \ref{sec:propagation}, TimeCycle, UVC, CorrFlow, and MAST apply the ``per-frame'' approach where a separate affinity matrix is computed for each context frame, while CRW uses the ``overall'' approach where a single affinity matrix is computed across concatenated features for all context frames. These two variants also differ in how softmax is applied to affinities, as explained in Sec. \ref{sec:propagation}.
Figure \ref{table:hyperparameter-studies} plots DAVIS performance for overall and per-frame propagation methods for a number of models as a function of other important hyper-parameters, namely softmax temperature, number of context frames, and number $k$ of top neighbors used for label transfer.
For the most part, the overall method outperforms per-frame for ImageNet, Rotation, MoCo, and CRW, while per-frame is best for UVC, although the performance of each method can vary widely based on hyperparameter choices, so each method must be further carefully tuned for fair comparisons.

\textbf{Temperature.} 
Figure \ref{table:hyperparameter-studies} (a) shows the effect of temperature $T$ in the affinity softmax normalization. While $T=1$ consistently works well across models in the ``overall'' setting, the best value of $T$ (and resulting performance) differs considerably across models in the ``per-frame'' approach. Accordingly, in the subsequent experiments for all datasets, we use the best temperature settings found here: 40 for ImageNet, 40 for Rotation, 50 for MoCo, 1 for UVC, and 50 for CRW.

\textbf{Number of context frames $n$.} 
Figure \ref{table:hyperparameter-studies} (b) shows that the choice of the number of context frames $n$ can also lead to meaningful performance differences. For ``overall,'' accuracy tends to increase monotonically with the number of frames, while for ``per-frame'' it typically peaks at intermediate values. Previous works have adopted different settings for $n$ with little discussion or comparison. For example, Jabri et al. \cite{jabri2020walk} report results with 20 context frames along with the first sequence frame while MAST uses 5 context frames including the first and fifth frame and CorrFlow uses only a single prior frame as context. In our main experiments (Section \ref{sec:davis-comp}), we will provide both results from a fixed setting similar to TimeCycle and UVC with 8 context frames (including the first frame in the sequence) and settings with additional context frames. For experiments in this work, we include only previous frames when discussing number of context frames $n$ and assume the first frame is also used as context unless otherwise stated.

\textbf{Number of neighbors $k$.} Figure \ref{table:hyperparameter-studies} (c) examines the effect of the $k$ parameter. We find that performance of all methods tends to be very sensitive to $k$ in the ``overall'' approach and a low value around 5 is best in most cases. By contrast, the ``per-frame'' approach is much less sensitive. Further, the trend for UVC tends to be different from the other methods' in both cases, as its performance seems to improve with larger values of $k$. Just as with $n$ above, different works used different values of $k$ with no explanation. Specifically, TimeCycle and UVC mostly used $k=5$ in their reported experiments, though UVC showed further improvement with $k=50$ for the VIP dataset in their released implementation, and CRW used $k=10$ (our Figure \ref{table:hyperparameter-studies} (c) indicates that $k=5$ may be slightly better, but it is produced without spatial localization, which gives further improvements as discussed next).

\setlength{\tabcolsep}{4pt}
\begin{table}
\begin{center}
\caption{Performance of selected methods on DAVIS 2017 validation dataset using spatial context mechanisms. All image-based models have been finetuned with modified stride as described in Sec. \ref{sec:implementation-facts:feat-extract}. All results in this table use context frames $n=7$ and $k=5$ neighbors during evaluation. Across image-based models, using the overall aggregation approach with fixed region localization consistently performs the best.
}
\label{table:davis-loc-studies}
{\small
\begin{tabular}
{p{0.025\textwidth}p{0.1\textwidth}p{0.11\textwidth}>{\centering}p{0.08\textwidth}>{\centering\arraybackslash}p{0.08\textwidth}}
\toprule
 &          &               &  \multicolumn{2}{c}{$\mathcal{J}\&\mathcal{F}_{\mathcal{M}}$} \\ 
\cmidrule(lr){4-5}
& Method & Localization & Per-frame & Overall                     \\
\midrule
(a) & Rotation & None         & 51.3      & 57.4                        \\
    &          & Fixed region & 58.6      & \textbf{62.3}                        \\
    &          & Track        & 54.1      & \textbf{-} \\
\midrule
(b) & MoCo     & None          & 55.6      & 60.0                        \\
    &          & Fixed region & 58.9      & \textbf{62.2}                        \\
    &          & Track        & 59.4      & \textbf{-} \\
\midrule
(c) & ImageNet & None         & 56.4      & 61.8                        \\
    &          & Fixed region & 60.4      & \textbf{65.1}                        \\
    &          & Track        & 59.3      & \textbf{-} \\
\midrule
(d) & UVC & None         & 59.4  & 49.0 \\
    &     & Fixed region & 56.5 & 48.2  \\
    &     & Track        & \textbf{60.8} & \textbf{-} \\
\midrule
(e) & CRW & None         & 56.2 & 62.1 \\
    &     & Fixed region & 59.0 & \textbf{65.8} \\
    &     & Track        & 60.3 & \textbf{-}    \\
\bottomrule
\end{tabular}
}
\end{center}
\end{table}
\setlength{\tabcolsep}{1.4pt}

\textbf{Spatial context localization.} As mentioned in Section \ref{sec:propagation}, to improve performance, several previous methods have incorporated mechanisms for restricting propagation to a localized region. These mechanisms can be considered a form of affinity matrix masking as outlined in Algorithm 1 (line 23)\algoreference{}. In particular, as described in Sec. \ref{sec:correspondence_methods}, UVC introduces a ``track" method, which restricts propagated masks to an ROI bounding box estimated at each step. Two important facts must be noted about this ``track'' method: it is dependent on the object mask since it uses affinities for all predicted mask locations to compute the ROI dimensions; though it has no trainable parameters itself, it is incorporated into UVC's training pipeline in combination with ``per-frame'' affinity matrix computation. The dependence on object mask ROI also means this mechanism is only used on the DAVIS object mask propagation task. By contrast, CRW uses a simpler ``fixed region'' mechanism, which propagates labels only from context regions within a radius of a particular target location (and this mechanism is not included in CRW's training). Lastly, CorrFlow propagates labels only from a reduced-size ROI around a target location, and MAST adds a more sophisticated coarse matching for determining ROI over a longer term. Unfortunately, most of these works merely report their results with the best combination of settings, making it difficult to understand how the above propagation heuristics impact performance.

The varied and highly customized nature of the above-mentioned methods is too great to permit exhaustive comparison, but in Table \ref{table:davis-loc-studies}, we compare UVC's ``track'' method (using their released code) and CRW's ``fixed region'' method with a radius of 12. The ``track'' method is reported only for the per-frame approach due to its dependence on column-wise softmax normalization of the full affinity matrix for each context frame. In all cases, adding a localization mechanism (either ``fixed region'' or ``track'') improves performance over using none. For all image-based models, ``fixed region'' works better than ``track,'' and best performance is reached for ``overall'' affinity matrix computation combined with ``fixed region'' propagation. The behavior of the CRW method is similar to that of image-based models in that it strongly benefits from the combination of ``overall'' affinity matrix computation and ``fixed region'' propagation. On the other hand, the UVC model only benefits from the ``track" mechanism which is included as part of its training pipeline (see \ref{sec:correspondence_methods}). Since the UVC method alone includes both per-frame affinity matrix computation and tracking in its training process, we find that it is very sensitive to changes of these settings at test time. In the following section, we will report results both with and without localization for fair comparison.

\textbf{Strong image baselines.} 
Based on the ablation studies of this section, we can conclude which combination of settings gives the best performance to still image features. Namely, if we use the supervised ImageNet model with ``overall'' propagation, 20 context frames, and ``fixed region'' localization, we get 65.8  $\mathcal{J}\& \mathcal{F}_{\mathcal{M}}$. It is worth noting that this is higher than any other numbers for an ImageNet ResNet18 reported in previous literature (and earlier works have sometimes cited very suboptimal numbers, as shown in Figure 1). Specifically, Jabri et al.~\cite{jabri2020walk} and Xu and Wang~\cite{xu2021rethinking} report 62.9 $\mathcal{J}\& \mathcal{F}_{\mathcal{M}}$, which is obtained without stride fine-tuning. The 3-point improvement of our ImageNet baseline matters because the amounts by which published methods claim to exceed the state of the art (or some particular baselines) are often smaller than that, and because researchers often go to considerable lengths to fine-tune their own methods, while paying less attention to similarly improving the baseline. Similarly, for MoCo, our best result is 62.7 $\mathcal{J}\& \mathcal{F}_{\mathcal{M}}$, higher than the 60.8 of~\cite{jabri2020walk,xu2021rethinking}. In Section \ref{sec:davis-comp}, we will revisit comparisons of recently introduced methods with the strong image-based baselines in mind.

\section{Combining image and video losses: UVC+}\label{sec:davis-comp:uvc-studies}

\setlength{\tabcolsep}{4pt}
\begin{table*}
\begin{center}
\caption{Performance on DAVIS 2017 validation dataset of the UVC+ framework which incorporates UVC \cite{li2019joint} with image-based visual representation learning. We study using
ImageNet classification and MoCo training as alternate visual representation learning signals to the Rotation \cite{gidaris2018unsupervised} task we use in our main results. 
We compare with an improved UVC model which uses the same hyperparameters as our models.
Under color, \cmark indicates whether the model is trained with any RGB images.
}
\label{table:davis-uvcplus}
{\small
\begin{tabular}
{p{0.12\textwidth}>{\centering}p{0.175\textwidth}>{\centering}p{0.18\textwidth}>{\centering}p{0.11\textwidth}>{\centering}p{0.07\textwidth}>{\centering}p{0.05\textwidth}>{\centering}p{0.05\textwidth}>{\centering}p{0.05\textwidth}>{\centering\arraybackslash}p{0.05\textwidth}}
\toprule
Method           & Auxiliary Loss          & Training Data    & Color  & $\mathcal{J}\&\mathcal{F}_{\mathcal{M}}$ &
$\mathcal{J}_{\mathcal{M}}$ & $\mathcal{J}_{\mathcal{O}}$ & $\mathcal{F}_{\mathcal{M}}$ & $\mathcal{F}_{\mathcal{O}}$ \\
\midrule
UVC \cite{li2019joint}    & \textbf{-}        & Kinetics    &  \textbf{-} &  57.8  & 56.3   & 65.0    & 59.2    & 64.1  \\
UVC (ours)        & \textbf{-}            & Kinetics          & \textbf{-} &  59.4  &  57.9                                          & 67.3                                          & 60.9                                          & 67.1                                          \\
UVC+       & Rotation & Kinetics          & \textbf{-} &  61.6  & 59.7  & 69.6  & 63.5  & 72.4                                          \\
\midrule 
UVC+ & Rotation        & Kinetics          & \cmark     & 62.6  & 60.6 & 71.6 & 64.6 & 74.8 \\
& Rotation  & Kinetics+ImageNet & \cmark & 62.3 & 60.3 & 70.2 & 64.3 & 72.4 \\
    & MoCo & Kinetics & \cmark & 61.0 & 58.8	& 67.4	& 63.2	& 70.0  \\
    & MoCo & Kinetics+ImageNet & \cmark & 60.8  &  58.5	& 67.2 & 	63.0 & 	69.7  \\
    & ImageNet       & Kinetics+ImageNet & \cmark   &  63.2  & 61.0 & 71.8 & 65.3   & 74.2                                          \\

\bottomrule 
\end{tabular}
}
\end{center}
\end{table*}
\setlength{\tabcolsep}{1.4pt}

The strong performance of image-based features, as discussed above, leads us to wonder whether video-based objectives can be augmented with image-based ones to further improve performance. To investigate this, we explore joint training of the UVC method on video and still image objectives.\footnote{We also considered adding an image-based objective to the CRW scheme. However, we found training of CRW from scratch to be prohibitively expensive and did not observe improvements from fine-tuning alone.} For the image-based auxiliary training, we explore the still image tasks introduced in Sec.~\ref{section:approaches:image}: rotation prediction, MoCo, and supervised ImageNet classification. 

To incorporate an image-based objective into the UVC framework, we add a subnetwork to produce the required outputs for each image-based task on top of the shared backbone network. 
During each training iteration, we sample a batch $\mathcal{V}$ of video data and a batch $\mathcal{I}$  of image data (or single video frames), and compute the UVC losses $\mathcal{L}_{V}$ for the video batch and image-based training losses $\mathcal{L}_{I}$ for the image batch before performing a joint optimization step on the combined loss:
$$\mathcal{L}(\mathcal{I},\mathcal{V}) = \sum_{V \in \mathcal{V}} \mathcal{L}_{V}(V) + \lambda \sum_{I \in \mathcal{I}} \mathcal{L}_{I}(I) \,, $$ 
\noindent where $\lambda$ is a hyperparameter to balance the two losses.

We refer to a model trained in the above manner as UVC+. In our implementation, we use the backbone architecture and hyperparameter settings consistent with the UVC model training described in Sec. \ref{sec:training_details}. Since the UVC backbone ResNet18 model is modified to have a large output grid dimension, we design the auxiliary subnetwork for image tasks to consist of a maxpooling layer, a residual block layer similar to the final ResNet18 layer, average pooling, and a final fully connected layer.
The video batch size is fixed at 40, as is used for UVC, while the image batch size is adjusted based on GPU memory consumption of image-based tasks: 64 for Rotation, 32 for MoCo, and 256 for ImageNet.
As for the tradeoff constant $\lambda$, we use $0.3$ for Rotation, $0.1$ for MoCo, and $0.05$ for ImageNet (these values were selected to give similar scales to the image and video losses early in training).

Table \ref{table:davis-uvcplus} shows the performance of the UVC+ method with each of the image-based learning tasks from Sec.~\ref{section:approaches:image}. For reference, the top line shows the reported UVC numbers from~\cite{li2019joint}. The second line, UVC (ours), shows the results from the model we trained from scratch using additional iterations as specified in Sec. \ref{sec:training_details}. The third line shows the result of our UVC+ model with the Rotation task trained on Kinetics frames converted to grayscale as in the original UVC approach, which involves a colorization loss. However, the UVC+ auxiliary loss is not colorization-based, so it allows for learning from RGB images or frames. Accordingly, the performance of UVC+ with the Rotation loss on color frames, shown in the fourth line, is about a point higher.

The bottom half of Table \ref{table:davis-uvcplus} compares performance of UVC+ with different auxiliary losses (Rotation, MoCo, ImageNet) computed on different data (either Kinetics frames, which are used for the UVC losses, or ImageNet training images, which are required for the supervised classification loss, and can optionally be used for the self-supervised ones).
Interestingly, while the MoCo model substantially outperformed the Rotation model for most settings in Table \ref{table:davis-loc-studies}, UVC+Rotation outperforms UVC+MoCo by over a point in both DAVIS metrics. Thus, for experiments in sections that follow, we use the Rotation task for UVC+ since it is the better performing self-supervised approach. UVC+ImageNet still beats UVC+Rotation, but just barely.

For the self-supervised UVC+MoCo and UVC+Rotation models, training only on Kinetics data rather than Kinetics+ImageNet does not significantly change results and in fact slightly improves performance in both cases. While ImageNet has $\sim$1.3M images compared with the $\sim$240k Kinetics-400 videos, each Kinetics video has several dozen individual frames from which we sample during training. We hypothesize that both the larger number of total frames in the Kinetics dataset and the smaller domain gap between Kinetics frames and the downstream video dataset outweigh the potential benefits from diversity of ImageNet image data. Similar trends hold across evaluation datasets.

We can summarize the take-aways of our UVC+ experiments as follows. First, self-supervised image-based losses do indeed hold the potential to improve over purely video-based losses for video correspondence. For UVC, the improvement from the previously published results in~\cite{li2019joint} to our best self-supervised model is 5.3 $\mathcal{J}$\&$\mathcal{F}$ mean. This makes the development of augmented objectives for other methods, like CRW\cite{jabri2020walk}, a promising future direction. Second, different image-based models may perform differently in combination with video-based losses than in isolation. Thus, when taken by itself, MoCo tended to outperform Rotation in Section \ref{sec:implementation-facts} (which is consistent with the generally reported superiority of contrastive models over older self-supervised models for still image tasks), but Rotation performed better in UVC+. Thus, it remains difficult to predict how well pre-trained models will transfer to substantially different tasks (in our case, from whole-image classification to video correspondence), and one should not always presume that the ``fancier'' recent models will be superior.

\section{Comparison of State-of-the-art Methods on DAVIS}\label{Comparative}\label{sec:davis-comp} \label{sec:davis-comp:analysis}

\begin{table*}
\begin{center}
\caption{Performance on DAVIS 2017 validation dataset of self-supervised temporal correspondence methods trained on video. 
Context denotes the number of context frames used while localization denotes if a type of spatial context localization is used. TimeCycle results are obtained with a $480\times480$ grid and a ResNet50 model (we also trained a ResNet18 model and evaluated at $480\times$full frames but obtained similar performance). 
Notations: \small{\emph{*Uses supervised labels. $\dagger$Result was obtained by evaluating the released model with corresponding code after removing localization and setting context frames to 8 for fair evaluation. **Unlike other methods, CorrFlow does not use initial frame in context. $\dagger\dagger$As described in Sec. \ref{sec:propagation}, CorrFlow and MAST use a modified per-frame approach with softmax normalization only over masked regions.}} }
\label{table:davis-main}
{\small
\begin{tabular}
{p{0.2\textwidth}>{\centering}p{0.125\textwidth}>{\centering}p{0.145\textwidth}>{\centering}p{0.125\textwidth}>{\centering}p{0.075\textwidth}>{\centering}p{0.07\textwidth}>{\centering}p{0.05\textwidth}>{\centering}p{0.05\textwidth}>{\centering}p{0.05\textwidth}>{\centering\arraybackslash}p{0.05\textwidth}}
\toprule
Method &  Training Data &  Prop. Approach & Localization & Context & $\mathcal{J}\&\mathcal{F}_{\mathcal{M}}$ & $\mathcal{J}_{\mathcal{M}}$ & $\mathcal{J}_{\mathcal{O}}$ & $\mathcal{F}_{\mathcal{M}}$ & $\mathcal{F}_{\mathcal{O}}$ \\
\midrule && \multicolumn{3}{c}{(a) No localization, fixed context, grid size $1\times$} &&&&&\\
\cmidrule(lr){3-5}
 MoCo     & ImageNet & overall & \textbf{-}    & 7   & 60.0 & 58.1 & 66.9 & 61.9 & 70.7 \\
ImageNet* & ImageNet & overall & \textbf{-} & 7    & 61.8 & 59.4 & 68.7 & 64.2 & 75.0 \\

 TimeCycle \cite{wang2019learning} (480$\times$480) &   VLOG & per-frame & \textbf{-}  & 7 & 48.2 & 46.4 & 50.0 & 50.0 & 48.0 \\
 UVC \cite{li2019joint} & Kinetics & per-frame & \textbf{-} & 7  & 57.8 & 56.3 & 65.0 & 59.2 & 64.1 \\
 UVC+ & Kinetics & per-frame & \textbf{-} & 7 & \textbf{62.6} & 60.6 & 71.6 & 64.6 & 74.8 \\
 CRW$\dagger$ \cite{jabri2020walk} & Kinetics & overall & \textbf{-} & 7 & 61.6 & 59.6 & 69.0 & 63.5 & 73.4 \\
\midrule
 && \multicolumn{3}{c}{(b) Localization, additional context, grid size $1\times$} &&&&&\\
\cmidrule(lr){3-5}

MoCo                    & ImageNet & overall & fixed region & 20 & 62.7 & 60.8 & 71.6 & 64.6 & 73.5 \\
ImageNet*   & ImageNet & overall & fixed region & 20 & 65.8 & 63.5 & 75.5 & 68.1 & 80.6 \\
 UVC \cite{li2019joint}  &  Kinetics & per-frame & track & 7  & 59.5 & 57.7 & 68.3 & 61.3 & 69.8 \\
 UVC+    & Kinetics & per-frame & track & 10 & 66.0 & 63.8 & 76.9  & 68.2 & 80.1 \\
 CRW \cite{jabri2020walk} & Kinetics & overall & fixed region & 20 & \textbf{67.5} & 64.8 & 76.1 & 70.2 & 82.1 \\
 VFS \cite{xu2021rethinking} & Kinetics & overall & fixed region    & 20 & 66.7 & 64.0 & - & 69.4 & - \\

\midrule && \multicolumn{3}{c}{(c) Localization, additional context, grid size $2\times$} &&&&&\\
\cmidrule(lr){3-5}
MoCo upsample & ImageNet & overall & fixed region & 20 & 65.8 & 63.2 & 73.6 & 68.3 & 79.1 \\
ImageNet upsample* & ImageNet & overall & fixed region &  20 & 67.8 & 65.1 & 75.0 & 70.5 & 81.6 \\
 CorrFlow** \cite{lai2019self} & OxUvA          &  per-frame$\dagger\dagger$    & fixed region    & 1  & 50.3 & 48.4    & 53.2      & 52.2   & 56.0   \\
 MAST \cite{lai2020mast} & YouTubeVOS     & per-frame$\dagger\dagger$      & predicted region    & 4  & 65.5 & 63.3    & 73.2    & 67.6    & 77.7     \\

 UVC+ upsample           & Kinetics          & per-frame        & track     & 10 & 67.6 & 65.0 & 75.7 & 70.1 & 79.6 \\
 CRW upsample & Kinetics & overall & fixed region & 20 & \textbf{70.6} & 68.5  &  79.3 & 72.7  & 85.8 \\

\bottomrule
\end{tabular}
}
\end{center}
\end{table*}

Informed by our hyperparameter studies, we would now like to compare top-performing self-supervised temporal correspondence methods in the literature. One possible way to get a fair comparison is to keep all settings the same across methods as much as possible, and another is to provide each method with the combination of settings that gives it the best performance. Needless to say, these two protocols might lead to different conclusions. For a full picture, we present a sequence of comparisons in Table \ref{table:davis-main}.

Table \ref{table:davis-main} (a) presents a comparison under a similar, minimalistic setting for all methods. Namely, we use no spatial context localization and keep the number of context frames fixed to 7, which is consistent with the protocols of TimeCycle and UVC. However, we keep the best-performing affinity computation scheme for each method, as indicated in the table. The table also notes that some video methods are trained on different datasets. In this restricted setting, UVC+ has a slight advantage, and both UVC+ and CRW barely outperform our strong ImageNet baseline (while the reported UVC result from~\cite{li2019joint} does not).

Next, in Table \ref{table:davis-main} (b), we compare the best result for each method by adding its preferred localization mechanism (``track'' for UVC and UVC+, and ``fixed region'' for CRW) and additional context frames (from Figure \ref{table:hyperparameter-studies} (b), we can see that UVC performance peaks at around 10 frames, while CRW keeps increasing as far as 20). In this regime, CRW has a slight advantage overall, and UVC+ and VFS are not far behind. Once again, no video-based method has a decisive advantage over ImageNet. The similar levels of performance are interesting, since the video-based methods exploit different combinations of cues: UVC+ uses a combination of cycle-consistency, colorization, and rotation losses; CRW uses primarily cycle-consistency and no colorization; and VFS uses similarity between nearby video frames and no localized cycle-consistency or colorization cues. The exact cues and losses do not seem to make much difference once we standardize the backbone and use a favorable combination of settings for each method.

Finally, Table \ref{table:davis-main} (c) performs comparisons with a larger output grid size, which was primarily used by CorrFlow and MAST. Despite its advantage in resolution, MAST is outperformed by CRW, as reported in Table \ref{table:davis-main} (b) above (and in the original CRW paper~\cite{jabri2020walk}). Still, to better understand the effects of grid size on accuracy, we evaluate UVC+, CRW, and the image-based models with a larger output grid. To do this without retraining, we simply upsample the input frames by a factor of 2 during inference. 
Even though the upsampled input to the trained models is blurry, and these networks are only trained on smaller images, we find that this change still significantly improves performance, as shown in Table \ref{table:davis-main} (c). CRW and MoCo find the largest improvements of 3.1 points $\mathcal{J}\&\mathcal{F}$ mean each, while UVC+ and supervised ImageNet improve by margins of 1.6 and 2.0 points respectively.
To our knowledge, the 70.6 $\mathcal{J}\&\mathcal{F}$ mean of upsampled CRW is the strongest result for self-supervised methods reported to date. We conjecture that this accuracy could be further improved by retraining CRW with the higher grid resolution and augmenting it with a still image objective similar either to UVC+ or to VFS. However, training such a model would be challenging due to the high computational intensity of CRW's dense patch-based extraction and similarity computations at high resolutions. 
A very recent work by Araslanov et al. \cite{araslanov2021dense} reaches comparable performance to CRW on DAVIS by exploring a related but much more efficient anchor-based dense self-supervised spatio-temporal learning scheme and could provide an interesting direction for future study on this topic.

\begin{figure*}
\begin{center}
\caption{Evaluation of the VIP semantic task using varied hyperparameter settings for selected models. Similarly to the DAVIS study in Table \ref{table:hyperparameter-studies},
we include study of context and nearest neighbors. Results are reported using the standard mIoU metric for the semantic task. We fix number of nearest neighbors $k=10$ in plot (a) and number of context frames $n=1$ in plot (b). Compared with DAVIS, additional context frames are less beneficial, and differences are less significant between per-frame and overall approaches. 
}
\label{table:vip-studies}
\includegraphics[width = 7.25in]{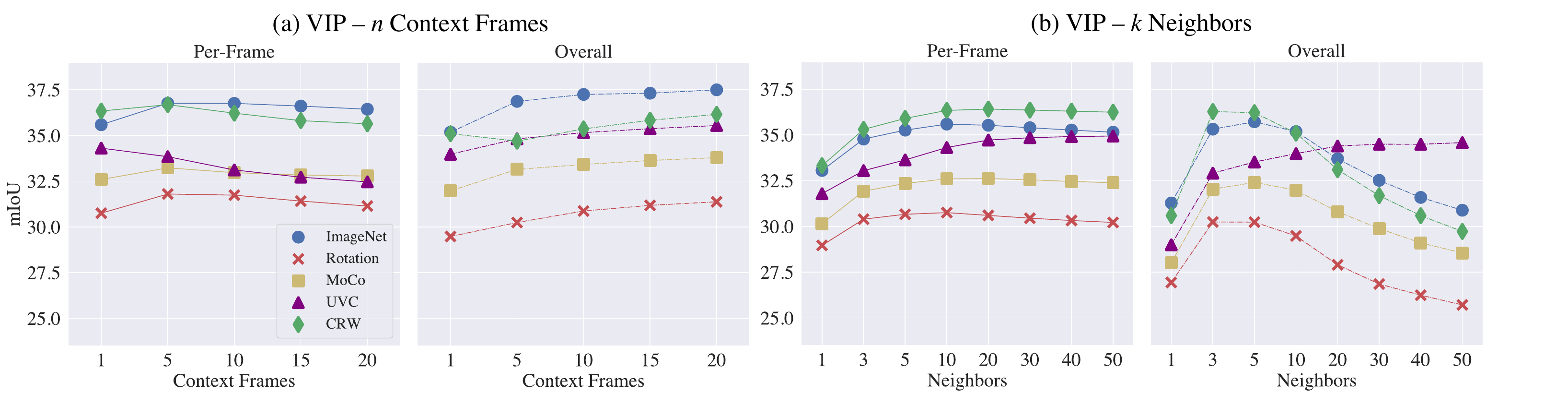}
\end{center}
\end{figure*}

\textbf{Fully supervised methods and online adaptation.} In absolute terms, the best results on the DAVIS task are obtained {\em fully supervised} methods trained on the DAVIS training set with  reference mask annotations. To our knowledge, the highest accuracy to date is 84.9 $\mathcal{J}\&\mathcal{F}_{\mathcal{M}}$, achieved by the AOT~\cite{yang2021associating} approach which uses a ResNet50 backbone network and transformer-based object association module capable of simultaneous matching and decoding of multiple object masks.
The CRW+upsample result lags around 14 points behind this. It is also worth noting that many recent supervised video object segmentation works perform test-time training, or online adaptation on each video during evaluation to improve performance \cite{caelles2017one,voigtlaender2017online,luiten2018premvos}. Jabri et al.~\cite{jabri2020walk} showed that self-supervised methods can also benefit from using these techniques, however, the improvements observed were under 1 point $\mathcal{J}\& \mathcal{F}_{\mathcal{M}}$ on DAVIS. Therefore, our experiments do not include test-time adaptation.

\section{Keypoint and Semantic Parts Propagation}\label{sec:vip-jhmdb}

Because the impact of implementation choices may be dataset-specific as well as method-specific, we extend our evaluation to the VIP and JHMDB datasets, which are common benchmarks for propagation of semantic parts and keypoints.

\vspace{0.5em}
\noindent\textbf{VIP.} The Video Instance-level Parsing (VIP) dataset~\cite{zhou2018adaptive} contains object mask labels similar to the DAVIS dataset. However, in this case, the object masks correspond to human parts (e.g., arm, leg, face). The dataset contains both semantic masks and instance-specific masks. 
Consistent with previous methods~\cite{wang2019learning,li2019joint,jabri2020walk}, we evaluate on the 50 videos in the validation set and resize input frames to 560$\times$560. Notably, the frames used for propagating labels in this dataset are significantly further apart than other datasets since only every 25th frame is annotated with dense labels. We evaluate semantic masks using the mean IoU metric and evaluate instance-level masks using mean average precision on the standard instance-level human parsing metric~\cite{li2017holistic}.

\setlength{\tabcolsep}{4pt}
\begin{table}
\begin{center}
\caption{Performance of methods on VIP dataset. 
Note that number of context frames includes the first frame which is not included in context frame count for some previous papers. Notations: \small{\emph{*Uses supervised labels. $\dagger$Results generated using released codebase.}}
}
\label{table:vip-main}
{\small
\begin{tabular}
{p{0.02\textwidth}p{0.12\textwidth}>{\centering}p{0.09\textwidth}>{\centering}p{0.06\textwidth}>{\centering}p{0.045\textwidth}>{\centering\arraybackslash}p{0.045\textwidth}}
\toprule
& Method & Localization & Context & mIoU & $AP^{r}_{vol}$ \\ 
\midrule 
\multirow{7}{*}{(a)} & MoCo & \textbf{-}& 1 & 32.6 & 16.6 \\
& ImageNet* \cite{li2019joint}   &  \textbf{-} & 1 & 31.8 & 12.6 \\
& ImageNet*    &  \textbf{-} & 1 &  35.7 & 17.5 \\
& TimeCycle \cite{wang2019learning} & \textbf{-} & 1 & 28.9 & 15.6 \\
& CRW$\dagger$ \cite{jabri2020walk} & \textbf{-} & 1 & 35.1 & - \\
& UVC \cite{li2019joint} & \textbf{-} & 1 & 34.1 & 17.7 \\
& UVC+ & \textbf{-} & 1  & \textbf{38.3}  & \textbf{22.2} \\
\midrule  
\multirow{4}{*}{(b)} & ImageNet* & fixed region & 4 & 37.7 & - \\
& UVC+ & fixed region & 4 & 38.9 & - \\
& CRW \cite{jabri2020walk} & fixed region & 4 & 38.6  & - \\
& VFS \cite{xu2021rethinking} & fixed region & 8 &  \textbf{39.9} & -\\

\bottomrule
\end{tabular}
}
\end{center}
\end{table}
\setlength{\tabcolsep}{1.4pt}

\setlength{\tabcolsep}{4pt}
\begin{table}
\begin{center}
\caption{Performance of methods on JHMDB dataset. 
Notations: \small{\emph{*Uses supervised labels. $\dagger$Results obtained using released CRW code and model which are lower than performance reported in paper.}}}
\label{table:jhmdb-main}
{\small
\begin{tabular}
{p{0.12\textwidth}>{\centering}p{0.09\textwidth}>{\centering}p{0.065\textwidth}>{\centering}p{0.06\textwidth}>{\centering\arraybackslash}p{0.06\textwidth}}
\toprule  
Method & Localization & Context & PCK@.1 & PCK@.2 \\
\midrule 
MoCo & \textbf{-} & 7 & 57.8 & 79.1 \\
ImageNet* \cite{li2019joint} & \textbf{-} & 7 &  53.8  & 74.6 \\
ImageNet*        & \textbf{-} & 7 & 59.2  & 80.4 \\ 
TimeCycle \cite{wang2019learning} & \textbf{-} & 7 & 57.3 & 78.1  \\
CorrFlow \cite{wang2019learning} & \textbf{-} & 7 & 58.5  & 78.8 \\ 
UVC\cite{li2019joint}  & \textbf{-} & 7 & 58.6      &   79.8    \\ 
UVC+       & \textbf{-} & 7 & 59.8   & \textbf{81.3} \\
CRW$\dagger$ \cite{jabri2020walk} & fixed region & 7 & 58.6 & 80.5 \\ 
VFS \cite{xu2021rethinking} & fixed region & 4 & \textbf{60.5} & 79.5  \\
\bottomrule
\end{tabular}
}
\end{center}
\end{table}
\setlength{\tabcolsep}{1.4pt}

Figure \ref{table:vip-studies} presents hyperparameter evaluations on the VIP dataset for different numbers $n$ of context frames and $k$ nearest neighbors, analogous to Figure \ref{table:hyperparameter-studies} in Section \ref{sec:implementation-facts}. 
Compared with the DAVIS studies of Figure \ref{table:hyperparameter-studies}, disparities between the per-frame and overall aggregation approaches are less prominent for the models evaluated. This could be partially due to the fact that methods do not benefit as significantly from additional context frames on VIP as shown in Figure \ref{table:vip-studies} (a). The limited benefit from additional context is likely because the VIP labels are included at sparse 25 frame intervals as mentioned above. This means that a longer context of 20 frames may average predictions over frames that are up to 20 seconds back which could introduce significant noise. 
A decreasing trend for adding more context frames is particularly prominent for UVC+ and CRW using the per-frame aggregation setting on VIP.

Table \ref{table:vip-main} contains full comparative results for both semantic and instance mask propagation on the VIP dataset. 
In Table \ref{table:vip-main} (a), we include experiments for the setting without spatial context localization and using only the initial frame and one previous frame as context. As with DAVIS, image-based models (MoCo, ImageNet) using stride fine-tuning perform very well compared with previously reported numbers. For example, the updated ImageNet classification model (third line) improves around +4 mIoU on the semantic task compared with previous ImageNet results reported by Li et al. \cite{li2019joint} (second line) and outperforms all video-based methods except for UVC+ under this setting.
Similarly to DAVIS, our UVC+ method performs very well in this setting, outperforming all other image-based and video-based methods for both the semantic task and instance task.

In Table \ref{table:vip-main} (b), we compare recent methods using spatial context localization and additional context frames on the semantic task. While CRW benefits substantially from spatial context localization and additional context frames, performance does not improve over UVC+. An ImageNet classification model performs slightly lower by around a point, while the recent VFS performs best by a 1 point margin. This increase may come from the additional context frames used by VFS, but other models do not generally gain more than 1 point in a similar setting as studied in Figure \ref{table:vip-studies}, implying strong performance of VFS.

Comparison to fully supervised models on VIP is not completely fair as standard supervised models do not take an initial frame label as input. However, it is still interesting to note that these self-supervised models compare quite favorably against the fully supervised ATEN~\cite{zhou2018adaptive} model which achieves 37.9 mIoU and 24.1 $AP^{r}_{vol}$ on the semantic and instance tasks respectively while relying on training from a large dataset of pixel-level annotations.

\vspace{0.5em}
\noindent\textbf{JHMDB.} 
The JHMDB dataset~\cite{jhuang2013towards} is a human actions dataset which contains human joint annotations. Each frame is annotated with 15 joint position labels. As in previous works, to propagate the keypoint labels we create a segmentation mask with each keypoint represented as a different class. We evaluate on test split 1 of the dataset,\footnotemark{} and all frames are resized to 240$\times$240 for evaluation. We evaluate predictions using the percentage of correct keypoints (PCK) metric~\cite{yang2012articulated} which considers keypoint coordinates to be correct when they are within a threshold of ground truth coordinates.

\footnotetext{This split used by self-supervised correspondence works is one of three standard splits over which results reported by supervised works are typically averaged. 
Prior works including TimeCycle~\cite{wang2019learning} and UVC~\cite{li2019joint} have also in some cases mistakenly compared to supervised results computed on a separate sub-JHMDB split containing only videos with fully visible keypoints.
}

Table \ref{table:jhmdb-main} shows comparative results from recent methods on JHMDB. 
We note once again that ImageNet model with stride fine-tuning (third line) performs very well compared with previous ImageNet classification results reported by Li et al. \cite{li2019joint} (second line). Self-supervised image-based models are on par with methods like TimeCycle and CorrFlow while the ImageNet classification model is comparable to top performing video-based methods. 
Though trends are similar to DAVIS and VIP on this dataset, we find that improvements are quite small between different methods. Further, there are not substantial performance increases from recent propagation improvements such as using spatial context localization. We therefore do not categorize results by these different implementation factor settings. 
For fully supervised comparison, previous models have achieved 68.7 PCK@0.1 and 81.6 PCK@0.2~\cite{song2017thin}, though these works evaluate slightly differently by averaging results across all three dataset splits.

\section{Conclusion}\label{sec:conclusion}
In this work, we carried out a systematic study of self-supervised temporal correspondence methods, with particular attention to implementation settings for feature extraction and label propagation. Much of the work we surveyed has suffered from several methodological flaws, notably (1) tuning hyperparameters directly on the test set, (2) under-tuning still image baselines, and (3) downplaying the importance of feature extraction and label propagation heuristics for achieving best performance. We hope that our study will better inform future research directions in temporal correspondence and help to improve evaluation protocols by showing the critical effects of certain implementation factors.

We found that well-optimized still image baselines perform better than previously reported, and in fact, no existing video-based method consistently and convincingly outperforms a fully supervised ImageNet baseline across datasets.
Interestingly, despite being based on different combinations of cues or losses, recent high-performing methods like CRW, VFS, and our augmented UCV+ method reach very similar levels of performance, close to the well-tuned ImageNet baseline, leading one to wonder how much these losses matter compared to more mundane engineering choices. In particular, VFS uses no tracking-based loss at all, only global similarity between video frames, suggesting that whatever the potential of localized correspondence cues like colorization or cycle consistency, it has not yet been fully realized. Developing a self-supervised video correspondence method capable of taking advantage of all the still image and video cues that have been identified remains an important open research direction. For this, it may be interesting to investigate recently proposed methods for dense self-supervised learning~\cite{wang21dense}. Another promising direction is to replace ad-hoc implementation choices in affinity matrix computation and label propagation by cleaner transformer-like attention architectures, similar to those recently proposed for video object segmentation~\cite{yang2021associating}.

{\small
\bibliographystyle{ieee_fullname}
\bibliography{egbib}

\begin{thebibliography}{10}\itemsep=-1pt

\bibitem{araslanov2021dense}
Nikita Araslanov, Simone Schaub-Meyer, and Stefan Roth.
\newblock Dense unsupervised learning for video segmentation.
\newblock In {\em Thirty-Fifth Conference on Neural Information Processing
  Systems}, 2021.

\bibitem{bao2018cnn}
Linchao Bao, Baoyuan Wu, and Wei Liu.
\newblock Cnn in mrf: Video object segmentation via inference in a cnn-based
  higher-order spatio-temporal mrf.
\newblock In {\em Proceedings of the IEEE Conference on Computer Vision and
  Pattern Recognition}, pages 5977--5986, 2018.

\bibitem{caelles2017one}
Sergi Caelles, Kevis-Kokitsi Maninis, Jordi Pont-Tuset, Laura Leal-Taix{\'e},
  Daniel Cremers, and Luc Van~Gool.
\newblock One-shot video object segmentation.
\newblock In {\em Proceedings of the IEEE conference on computer vision and
  pattern recognition}, pages 221--230, 2017.

\bibitem{caron2018deep}
Mathilde Caron, Piotr Bojanowski, Armand Joulin, and Matthijs Douze.
\newblock Deep clustering for unsupervised learning of visual features.
\newblock In {\em Proceedings of the European Conference on Computer Vision
  (ECCV)}, pages 132--149, 2018.

\bibitem{chen2020simple}
Ting Chen, Simon Kornblith, Mohammad Norouzi, and Geoffrey Hinton.
\newblock A simple framework for contrastive learning of visual
  representations.
\newblock In {\em International conference on machine learning}, pages
  1597--1607. PMLR, 2020.

\bibitem{chen2019self}
Ting Chen, Xiaohua Zhai, Marvin Ritter, Mario Lucic, and Neil Houlsby.
\newblock Self-supervised gans via auxiliary rotation loss.
\newblock In {\em Proceedings of the IEEE Conference on Computer Vision and
  Pattern Recognition}, pages 12154--12163, 2019.

\bibitem{chen2021exploring}
Xinlei Chen and Kaiming He.
\newblock Exploring simple siamese representation learning.
\newblock In {\em Proceedings of the IEEE/CVF Conference on Computer Vision and
  Pattern Recognition}, pages 15750--15758, 2021.

\bibitem{deng2009imagenet}
Jia Deng, Wei Dong, Richard Socher, Li-Jia Li, Kai Li, and Li Fei-Fei.
\newblock Imagenet: A large-scale hierarchical image database.
\newblock In {\em 2009 IEEE conference on computer vision and pattern
  recognition}, pages 248--255. Ieee, 2009.

\bibitem{gidaris2019boosting}
Spyros Gidaris, Andrei Bursuc, Nikos Komodakis, Patrick P{\'e}rez, and Matthieu
  Cord.
\newblock Boosting few-shot visual learning with self-supervision.
\newblock In {\em Proceedings of the IEEE International Conference on Computer
  Vision}, pages 8059--8068, 2019.

\bibitem{gidaris2018unsupervised}
Spyros Gidaris, Praveer Singh, and Nikos Komodakis.
\newblock Unsupervised representation learning by predicting image rotations.
\newblock In {\em ICLR 2018}, 2018.

\bibitem{he2020momentum}
Kaiming He, Haoqi Fan, Yuxin Wu, Saining Xie, and Ross Girshick.
\newblock Momentum contrast for unsupervised visual representation learning.
\newblock In {\em Proceedings of the IEEE/CVF Conference on Computer Vision and
  Pattern Recognition}, pages 9729--9738, 2020.

\bibitem{jabri2020walk}
Allan Jabri, Andrew Owens, and Alexei~A Efros.
\newblock Space-time correspondence as a contrastive random walk.
\newblock {\em Advances in Neural Information Processing Systems}, 2020.

\bibitem{jaderberg2015spatial}
Max Jaderberg, Karen Simonyan, Andrew Zisserman, et~al.
\newblock Spatial transformer networks.
\newblock {\em Advances in neural information processing systems},
  28:2017--2025, 2015.

\bibitem{jhuang2013towards}
Hueihan Jhuang, Juergen Gall, Silvia Zuffi, Cordelia Schmid, and Michael~J
  Black.
\newblock Towards understanding action recognition.
\newblock In {\em Proceedings of the IEEE international conference on computer
  vision}, pages 3192--3199, 2013.

\bibitem{kay2017kinetics}
Will Kay, Joao Carreira, Karen Simonyan, Brian Zhang, Chloe Hillier, Sudheendra
  Vijayanarasimhan, Fabio Viola, Tim Green, Trevor Back, Paul Natsev, et~al.
\newblock The kinetics human action video dataset.
\newblock {\em arXiv preprint arXiv:1705.06950}, 2017.

\bibitem{kolesnikov2019revisiting}
Alexander Kolesnikov, Xiaohua Zhai, and Lucas Beyer.
\newblock Revisiting self-supervised visual representation learning.
\newblock {\em arXiv preprint arXiv:1901.09005}, 2019.

\bibitem{lai2020mast}
Zihang Lai, Erika Lu, and Weidi Xie.
\newblock Mast: A memory-augmented self-supervised tracker.
\newblock In {\em Proceedings of the IEEE/CVF Conference on Computer Vision and
  Pattern Recognition}, pages 6479--6488, 2020.

\bibitem{lai2019self}
Zihang Lai and Weidi Xie.
\newblock Self-supervised learning for video correspondence flow.
\newblock {\em arXiv preprint arXiv:1905.00875}, 2019.

\bibitem{li2017holistic}
Qizhu Li, Anurag Arnab, and Philip~HS Torr.
\newblock Holistic, instance-level human parsing.
\newblock {\em arXiv preprint arXiv:1709.03612}, 2017.

\bibitem{li2019joint}
Xueting Li, Sifei Liu, Shalini~De Mello, Xiaolong Wang, Jan Kautz, and
  Ming-Hsuan Yang.
\newblock Joint-task self-supervised learning for temporal correspondence.
\newblock In {\em NeurIPS}, 2019.

\bibitem{long2014convnets}
Jonathan~L Long, Ning Zhang, and Trevor Darrell.
\newblock Do convnets learn correspondence?
\newblock In {\em Advances in neural information processing systems}, pages
  1601--1609, 2014.

\bibitem{luiten2018premvos}
Jonathon Luiten, Paul Voigtlaender, and Bastian Leibe.
\newblock Premvos: Proposal-generation, refinement and merging for video object
  segmentation.
\newblock In {\em Asian Conference on Computer Vision}, pages 565--580.
  Springer, 2018.

\bibitem{misra2020self}
Ishan Misra and Laurens van~der Maaten.
\newblock Self-supervised learning of pretext-invariant representations.
\newblock In {\em Proceedings of the IEEE/CVF Conference on Computer Vision and
  Pattern Recognition}, pages 6707--6717, 2020.

\bibitem{musgrave2020metric}
Kevin Musgrave, Serge Belongie, and Ser-Nam Lim.
\newblock A metric learning reality check.
\newblock In {\em European Conference on Computer Vision}, pages 681--699.
  Springer, 2020.

\bibitem{paszke2019pytorch}
Adam Paszke, Sam Gross, Francisco Massa, Adam Lerer, James Bradbury, Gregory
  Chanan, Trevor Killeen, Zeming Lin, Natalia Gimelshein, Luca Antiga, et~al.
\newblock Pytorch: An imperative style, high-performance deep learning library.
\newblock In {\em Advances in Neural Information Processing Systems}, pages
  8024--8035, 2019.

\bibitem{perazzi2016benchmark}
Federico Perazzi, Jordi Pont-Tuset, Brian McWilliams, Luc Van~Gool, Markus
  Gross, and Alexander Sorkine-Hornung.
\newblock A benchmark dataset and evaluation methodology for video object
  segmentation.
\newblock In {\em Proceedings of the IEEE Conference on Computer Vision and
  Pattern Recognition}, pages 724--732, 2016.

\bibitem{pont20172017}
Jordi Pont-Tuset, Federico Perazzi, Sergi Caelles, Pablo Arbel{\'a}ez, Alex
  Sorkine-Hornung, and Luc Van~Gool.
\newblock The 2017 davis challenge on video object segmentation.
\newblock {\em arXiv preprint arXiv:1704.00675}, 2017.

\bibitem{song2017thin}
Jie Song, Limin Wang, Luc Van~Gool, and Otmar Hilliges.
\newblock Thin-slicing network: A deep structured model for pose estimation in
  videos.
\newblock In {\em Proceedings of the IEEE conference on computer vision and
  pattern recognition}, pages 4220--4229, 2017.

\bibitem{voigtlaender2017online}
Paul Voigtlaender and Bastian Leibe.
\newblock Online adaptation of convolutional neural networks for video object
  segmentation.
\newblock {\em arXiv preprint arXiv:1706.09364}, 2017.

\bibitem{vondrick2018tracking}
Carl Vondrick, Abhinav Shrivastava, Alireza Fathi, Sergio Guadarrama, and Kevin
  Murphy.
\newblock Tracking emerges by colorizing videos.
\newblock In {\em Proceedings of the European Conference on Computer Vision
  (ECCV)}, pages 391--408, 2018.

\bibitem{wang2019learning}
Xiaolong Wang, Allan Jabri, and Alexei~A Efros.
\newblock Learning correspondence from the cycle-consistency of time.
\newblock In {\em Proceedings of the IEEE Conference on Computer Vision and
  Pattern Recognition}, pages 2566--2576, 2019.

\bibitem{wang21dense}
Xinlong Wang, Rufeng Zhang, Chunhua Shen, Tao Kong, and Lei Li.
\newblock Dense contrastive learning for self-supervised visual pre-training.
\newblock In {\em CVPR}, 2021.

\bibitem{wang2021different}
Zhongdao Wang, Hengshuang Zhao, Ya-Li Li, Shengjin Wang, Philip Torr, and Luca
  Bertinetto.
\newblock Do different tracking tasks require different appearance models?
\newblock In {\em Thirty-Fifth Conference on Neural Information Processing
  Systems}, 2021.

\bibitem{xu2021rethinking}
Jiarui Xu and Xiaolong Wang.
\newblock Rethinking self-supervised correspondence learning: A video
  frame-level similarity perspective.
\newblock {\em arXiv preprint arXiv:2103.17263}, 2021.

\bibitem{yang2012articulated}
Yi Yang and Deva Ramanan.
\newblock Articulated human detection with flexible mixtures of parts.
\newblock {\em IEEE transactions on pattern analysis and machine intelligence},
  35(12):2878--2890, 2012.

\bibitem{yang2021associating}
Zongxin Yang, Yunchao Wei, and Yi Yang.
\newblock Associating objects with transformers for video object segmentation.
\newblock {\em arXiv preprint arXiv:2106.02638}, 2021.

\bibitem{DMMNet19}
Xiaohui Zeng, Renjie Liao, Li Gu, Yuwen Xiong, Sanja Fidler, and Raquel
  Urtasun.
\newblock Dmm-net: Differentiable mask-matching network for video instance
  segmentation.
\newblock In {\em ICCV}, 2019.

\bibitem{zhai2019s4l}
Xiaohua Zhai, Avital Oliver, Alexander Kolesnikov, and Lucas Beyer.
\newblock S4l: Self-supervised semi-supervised learning.
\newblock In {\em Proceedings of the IEEE international conference on computer
  vision}, pages 1476--1485, 2019.

\bibitem{zhou2018adaptive}
Qixian Zhou, Xiaodan Liang, Ke Gong, and Liang Lin.
\newblock Adaptive temporal encoding network for video instance-level human
  parsing.
\newblock In {\em Proceedings of the 26th ACM international conference on
  Multimedia}, pages 1527--1535, 2018.

\end{thebibliography}
}

\end{document}